\documentclass[lettersize,journal]{IEEEtran}
\usepackage{amsmath,amsfonts}
\usepackage{array}
\usepackage[caption=false,font=normalsize,labelfont=sf,textfont=sf]{subfig}
\usepackage{textcomp}
\usepackage{stfloats}
\usepackage{url}
\usepackage{verbatim}
\usepackage{graphicx}
\usepackage{cite}
\usepackage{orcidlink}

\usepackage{breqn}
\usepackage{algorithmicx}
\usepackage[ruled, lined, linesnumbered, longend]{algorithm2e}
\usepackage{multirow}
\usepackage{makecell} 
\usepackage{arydshln}
\SetKwBlock{Loop}{Loop}{end}
\SetKwBlock{Foreach}{Foreach}{end}
\newcommand{\nonl}{\renewcommand{\nl}{\let\nl\oldnl}}
\SetKwInput{KwInput}{Input}                
\SetKwInput{KwOutput}{Output}              

\begin{document}

\title{Negative Sampling in Knowledge Graph Representation Learning: A Review}

\author{Tiroshan Madushanka \orcidlink{0000-0001-6627-1477} and Ryutaro Ichise
}




\maketitle

\begin{abstract}
Knowledge Graph Representation Learning (KGRL), or Knowledge Graph Embedding (KGE), is essential for AI applications such as knowledge construction and information retrieval. These models encode entities and relations into lower-dimensional vectors, supporting tasks like link prediction and recommendation systems. Training KGE models relies on both positive and negative samples for effective learning, but generating high-quality negative samples from existing knowledge graphs is challenging. The quality of these samples significantly impacts the model’s accuracy.
This comprehensive survey paper systematically reviews various negative sampling (NS) methods and their contributions to the success of KGRL. Their respective advantages and disadvantages are outlined by categorizing existing NS methods into six distinct categories. Moreover, this survey identifies open research questions that serve as potential directions for future investigations. By offering a generalization and alignment of fundamental NS concepts, this survey provides valuable insights for designing effective NS methods in the context of KGRL and serves as a motivating force for further advancements in the field.
\end{abstract}


%
\section{Introduction}
\label{section:introduction}

\IEEEPARstart{K}nowledge Graphs (KGs) effectively represent structured data in the form of $(head, relation, tail)$ triplets. They are constructed from knowledge bases such as NELL~\cite{Carlson2010NELL}, Freebase~\cite{Bollacker2008Freebase}, DBpedia~\cite{Auer2007DBpediaAN}, WordNet~\cite{Miller1995WordNet}, and YAGO~\cite{Suchanek2007Yago}, and have been applied across domains such as question-answering~\cite{Xiao2019QA, Davide2016ExemplarQueries, Nararatwong2022FinQA}, recommendation systems~\cite{Xiang2019KGAT, Fuzheng2016KGRecommender, Hongwei2018DKN}, and information retrieval~\cite{Venkatesh2022ConversationalIR, Dietz2018UtilizingKG}. Despite the richness of these knowledge bases, KGs often remain incomplete due to the evolving nature of real-world facts. For example, Freebase lacks 71\% of certain data types, while DBpedia lacks 66\%~\cite{Dong2014Kvault, Krompas2015TypeNS}. This highlights the need for automated methods to infer missing knowledge and facts in KGs.

In recent years, Machine Learning (ML) has gained traction in addressing KG completion. However, the high dimensionality of KG data presents significant challenges in processing and analysis. To address these, knowledge graph embedding (KGE) techniques have been developed, which map entities and relations into a lower-dimensional vector space while preserving their semantic relationships. KGE methods have demonstrated success in tasks such as link prediction, entity recognition, relation extraction, and triplet classification. Training KGE models typically involves ranking observed (positive) instances higher than unobserved (negative) ones. Since knowledge bases predominantly consist of positive instances, generating negative instances becomes essential to help the model learn the underlying semantics. Crafting high-quality negatives is both challenging and crucial for improving KG embeddings. Consequently, negative sampling (NS) has become a critical element in knowledge representation learning, significantly influencing KGE model performance through effective negative selection.

\IEEEpubidadjcol This paper surveys the literature on negative sampling methods in Knowledge Graph Representation Learning (KGRL). It explores the historical evolution of negative sampling techniques and recent developments in generating high-quality negative samples. The paper provides a detailed comparison of negative sampling strategies, categorizing them into six distinct groups based on their architectures. Each category is analyzed for its strengths and limitations. Additionally, this work identifies open challenges in negative sampling and suggests avenues for future research. The key contributions of this paper are:

\begin{itemize}

\item \textbf{Comprehensive review}: This paper presents a comprehensive review of the historical background of negative sampling in KGRL, as well as the contemporary methodologies employed for generating negative samples.

\item \textbf{In-depth classification and novel grouping}: The paper thoroughly compares the research conducted on negative sampling strategies in the context of knowledge graph representation learning. It introduces six distinct categories that encompass various approaches, providing a comprehensive overview of the architecture underlying each negative sampling category. Furthermore, the advantages and disadvantages of the proposed categories are systematically discussed.

\item \textbf{Future research prospects}: The paper concludes by outlining several unresolved research challenges in the field of negative sampling and directs readers toward relevant studies exploring the application of negative sampling in KGRL.
\end{itemize}

The subsequent sections of this manuscript are arranged as follows. Section~\ref{section:overview} presents an extensive overview of knowledge representation learning, encompassing negative sampling, its historical background, influence, and dimensions. Additionally, this section explicates the notations and definitions pertinent to this research domain. In Section~\ref{section:negative_sampling}, we scrutinize negative sampling approaches based on the newly proposed grouping. Subsequently, in Section~\ref{section:conclusion}, we deliberate on potential directions for future research and conclude this paper.
%
%
\section{Overview}
\label{section:overview}
\subsection{Review Methodology}
\begin{figure}[]
\centering
    \includegraphics[width=0.44 \textwidth]{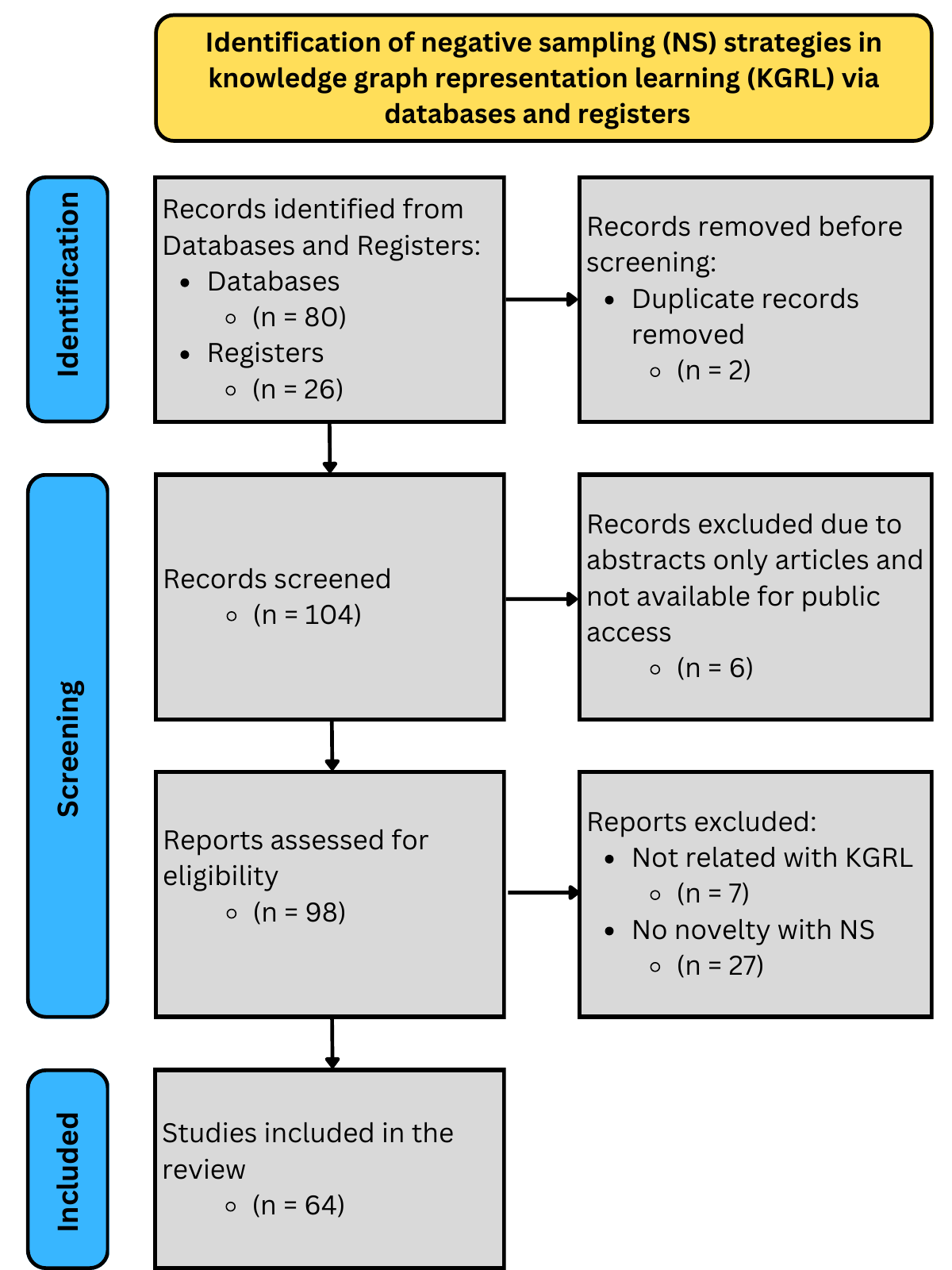}
\caption {PRISMA method used for selecting the articles in this survey.} 
\label{fig:flow}
\end{figure}

This systematic review was conducted following the PRISMA (Preferred Reporting Items for Systematic Reviews and Meta-Analyses) guidelines introduced by Moher et al.~\cite{Moher2009PRISM}. Relevant publications were meticulously selected using the process outlined below.

The selection process, illustrated in Figure~\ref{fig:flow}, details the steps taken to identify pertinent research papers. An extensive search was performed across several databases, including ACM Digital Library~\footnote{https://dl.acm.org/}, IEEE Xplore~\footnote{https://ieeexplore.ieee.org/}, ScienceDirect~\footnote{https://www.sciencedirect.com/}, Web of Science~\footnote{https://www.webofscience.com/}, and SpringerLink~\footnote{https://link.springer.com/}, identifying 80 research papers. Additionally, 26 papers were retrieved from the open-access repository ArXiv~\footnote{https://arxiv.org/}.

From the initial pool of 106 papers, 2 duplicates were removed. Furthermore, 6 articles were excluded due to being abstracts-only (1) or lacking public accessibility (5). This resulted in 98 articles for further review. A full-text analysis of these articles was conducted, after which 34 were excluded for being irrelevant to knowledge graph representation learning and negative sampling. Ultimately, 64 papers met the inclusion criteria and were incorporated into this systematic review.

\subsection{Definitions and Notations}

{\renewcommand{\arraystretch}{1.25}%
\begin{table}[t]
\caption{Notations and Description.}
\begin{center}
\resizebox{\columnwidth}{!}{
\setlength{\tabcolsep}{3pt}
\begin{tabular}{ll}
\Xhline{1pt} 
Notations & Description \\ 
\Xhline{1.3pt} 
$\mathcal{G}$         & A knowledge graph \\
$\mathcal{R}$         & Relation set \\
$\mathcal{E}$         & Entity set \\
$\mathcal{S}$         & Training triple set  \\
$(h, r, t)$ & A triple of head, relation, and tail \\
(\textbf{h}, \textbf{r}, \textbf{t}) & Embedding of the head, relation, and tail \\
$(\bar{h}, r, \bar{t})$ & A negative triple with corrupted head, relation, and tail \\
$n$ = $\mid \mathcal{S} \mid$   & Number of triples \\
$\gamma$         & Margin parameter \\
$f($·$)$     & Score function \\ 
M$_{r}$     & Mapping matrix \\
$I$         &  Identity matrix \\
$\mathbb{R}^d$  & $d$-dimensional real-valued space \\
$\mathbb{T}^d$  & $d$-dimensional torus space \\
$\mathbb{E}[X]$  & Expected value of random variable $X$ \\
\Xhline{1.3pt} 
\end{tabular}
}
\end{center}
\label{table:notation}
\end{table} 

We adopt the following key definitions and notations to facilitate a systematic comparison and analysis of negative sampling methods in KGRL. First, we follow Wang et al.'s definition of multi-relational knowledge graphs~\cite{Wang2021Survey}, labeled as Definition \ref{defn:kg}, which provides a foundational understanding of knowledge graph structures. Second, we incorporate Ji et al.'s definition of knowledge graph embedding~\cite{Ji2022Survey} (i.e., Definition \ref{defn:kge}), which outlines the principles and techniques of mapping knowledge graph entities and relations to low-dimensional spaces. Lastly, we reference the concept of Quality Negatives~\cite{tiroshan2022tuckerdncaching} introduced by Madushanka and Ichise, presented as Definition \ref{defn:qneg}, offering insights into the generation and use of high-quality negative samples. These definitions form the basis for our analysis of negative sampling techniques in knowledge graph representation learning.

\newtheorem{mydef}{Definition}
\begin{mydef}\mbox{}(\textbf{Knowledge Graph} )\label{defn:kg}
\begin{itemize}
\item[]
A knowledge graph is a multi-relational graph composed of entities and relations regarded as nodes and different types of edges, respectively.
\end{itemize}
\end{mydef}

\begin{mydef}\mbox{}(\textbf{Knowledge Graph Embedding)} )\label{defn:kge}
\begin{itemize}
\item[]
Knowledge graph embedding is a technology for mapping the content of entities and relations in a knowledge graph to continuous low-dimensional vector space.
\end{itemize}
\end{mydef}

\begin{mydef}\mbox{}(\textbf{Quality Negative Sample} )\label{defn:qneg}
\begin{itemize}
\item[]
A quality negative is a semantically meaningful but factually incorrect triplet that is hard to distinguish without referring to the ground truth. 
\end{itemize}
\end{mydef}

As detailed in Table~\ref{table:notation}, this study employs specific notations for clarity. A knowledge graph is denoted as $\mathcal{G} = (\mathcal{E}, \mathcal{R})$, where $\mathcal{E}$ represents the set of entities and $\mathcal{R}$ the set of relations. Head and tail entities are represented by $h$ and $t$, respectively, with relations denoted by $r$, which connect $h$ and $t$. A relation $r$ is expressed as $(h, t)$, where $h, t \in \mathcal{E}$, and $r \in \mathcal{R}$. Knowledge graph facts are represented as triplets $(h, r, t)$, with the training set $\mathcal{S}$ comprising a collection of such facts, i.e., $\mathcal{S} = \{(h, r, t)\}$. Negative head samples are represented by $\bar{h}$ and negative tail samples by $\bar{t}$. 

\subsection{Knowledge Graph Representation Learning}
\label{subsection:kgrl}

The challenge of knowledge graph completion has driven the development of methods to enhance machine understanding and reasoning over KGs. Knowledge graph representation learning, or knowledge graph embedding, has become particularly prominent among these. This method projects entities and relations from a KG into a low-dimensional continuous vector space, enabling efficient knowledge tasks. In KGE models, a scoring function evaluates entity interactions based on relations, often framed as an energy function~\cite{Ji2022Survey}. Energy-based learning learns an energy function $\mathbf{}{E}_\theta(x)$, where $\theta$ represents the parameters and $x$ is the input, ensuring positive samples score higher than negative ones.

KGRL models are typically divided into four main types, including two traditional and two newer approaches: geometric models, tensor decomposition models, deep learning models, and semantic-aware models. Geometric models treat relations as geometric transformations within a latent space, focusing on spatial relationships between entities. Tensor decomposition models use mathematical techniques to capture deeper semantic meanings. Deep learning models employ neural networks, extracting patterns through parameter learning, while semantic-aware models enrich embeddings by integrating additional information to preserve the semantic essence of entities and relations. These models aim to uncover the inherent connections and structures in the knowledge graph.

\subsubsection{Geometric Models} 

Geometric models interpret relations as geometric shifts in a latent space, using distance functions like \textbf{L1} or \textit{L2} norms to measure the proximity of entities. A well-known example is TransE~\cite{Bordes2013Translating}, which connects entity embeddings \textbf{h} and \textbf{t} with relation embedding $\textbf{r}$ for a positive triplet $(h,r,t)$. However, TransE struggles with complex relation types like one-to-many or many-to-many, leading to the development of models like TransH~\cite{Wang2014Knowledge}, TransR~\cite{Lin2015Learning}, and TransD~\cite{Ji2015Knowledge}, which extend embeddings into different spaces to model relational diversity better. These models differ in how they project entities and relations: TransH uses a hyperplane, TransR introduces separate spaces for entities and relations, and TransD employs dynamic mapping matrices for greater flexibility. TransHR~\cite{Zhang2017TransHR} extends this to hyper-relational data, while TransA~\cite{Xiao2015TransA} employs the Mahalanobis distance for adaptive metric learning, enhancing the model's ability to capture fine-grained relations. TransM~\cite{Fan2014TransitionbasedKG} introduces a weight factor for distance calculation, adapting to the varying distances between entities and relations.
TransF~\cite{Feng2016TransF} distinguishes itself with a dot product scoring function $f(h,r,t) = (\textbf{h} + \textbf{r})^\top \textbf{t}$, allowing more flexible translations in embedding space. ITransF~\cite{xie2017DomainNS} uncovers hidden concepts via sparse attention vectors, and TransAt~\cite{Qian2018TransAt} enhances learning with relation attention. TransMS~\cite{Yang2019TransMS} captures multi-directional semantics using nonlinear functions and linear bias vectors.

ManifoldE~\cite{Xiao2016ManifoldE} combines distance-based scoring with manifold structures, enhancing the geometry of embedding spaces. Models like KG2E~\cite{He2015K2GE} and TransG~\cite{Xiao2016Transg} leverage Gaussian distributions to capture uncertainties. KG2E uses Gaussian vectors to measure triple compatibility, while TransG employs a mixture of Gaussians to model semantic variability. HAKE~\cite{Zhang2020HAKE} offers a novel approach by representing entities in a hierarchical structure using polar coordinates. Entities are differentiated by their hierarchical level (modulus) and proximity (phase), capturing nuanced relationships.

Alternative models like TorusE~\cite{Takuma2018TorusE}, QuatE~\cite{Zhang2019QuatE}, RotatE~\cite{Sun2019RotatEKG}, and MobiusE~\cite{Chen2021MobiusE} use advanced mathematical concepts to model relations. Despite their differences, these methods share the underlying idea of mapping head entities to tail entities via relations. RotatE, for instance, models relations as rotations in complex vector space, while TorusE maps embeddings onto a torus to preserve geometric properties.

\subsubsection{Tensor Decomposition Models}

Tensor decomposition models extract latent semantics by decomposing the interactions between entities and relations. RESCAL~\cite{Nickel2011ATM}, one of the earliest examples, uses a matrix to model relationships between features of entity embeddings. Variations like DistMult~\cite{Yang2015EmbeddingEA}, HolE~\cite{Nickel2016HolographicEO}, and ComplEx~\cite{Trouillon2016ComplexEF} simplify or extend this approach. DistMult replaces the interaction matrix with a diagonal one, simplifying computations, while HolE uses a circular correlation operator to compress tensor products. ComplEx, working in complex space, captures more intricate relationships between entities.

SimplE~\cite{Kazemi2018SimplEEF} streamlines ComplEx, reducing model complexity while maintaining competitive performance. HolEx~\cite{Xue2018Holex} bridges HolE and tensor products, balancing efficiency and expressiveness. CrossE~\cite{Zhang2019CrossE} introduces crossover interactions, capturing bidirectional relationships between entities and relations. ANALOGY~\cite{Liu2017Analogical} focuses on multi-relational inference by modeling analogical structures, while TuckER~\cite{Balazevic2019TuckERTF} applies Tucker decomposition to capture more productive entity-relation interactions.

\subsubsection{Deep Learning Models}

Deep learning models use neural networks for Link Prediction, organizing neurons into layers, and learning patterns through weights and biases. ConvE~\cite{Dettmers2018Convolutional2K} reshapes entity and relation vectors into a matrix passed through convolutional layers, producing a final score via a dot product with the tail entity. ConvKB~\cite{Nguyen2018Convkb} concatenates vectors into a matrix for convolution, while ConvR~\cite{Jiang2019ConvR} applies relation-specific filters. InteractE~\cite{Shikhar2020interacte} enhances ConvE by stacking multiple permutations of input vectors, processed through circular convolution. CapsE~\cite{Nguyen2019CapsE} adds capsule layers, each capturing distinct features of the input facts.

\subsubsection{Semantic-Aware Models}
Semantic-aware models blend graph structure and natural language understanding. Traditional methods focus on local structure, while PLM-based approaches incorporate linguistic context from entity names and descriptions. KG-BERT~\cite{Yao2019KGBERTBF} was one of the first to use BERT~\cite{Devlin2019BERT} to extract features for link prediction. ERNIE~\cite{Zhang2019Ernie} extends this by integrating lexical, syntactic, and knowledge-based information. StAR~\cite{Bo2021Star} and SimKGC~\cite{Wang2022Simkgc} both use dual transformers to extract textual representations, while KEPLER~\cite{Wang2019KEPLERAU} and InductivE~\cite{Wang2021InductivE} combine textual features with traditional embeddings. BERTRL~\cite{Zha2022InductiveRP} fine-tunes PLMs with relation paths, while KGT5~\cite{saxena2022kgt5} pre-trains on KG tasks using a seq2seq transformer model.

\subsection{Negative Sampling}

The notion of negative sampling was first introduced as importance sampling in probabilistic neural models of language~\cite{Schroff2015FaceNet}. Noise Contrastive Estimation (NCE)~\cite{Gutmann2012NCE} was initially developed to overcome the computational difficulties of probabilistic language models, which involve assessing partition functions across an extensive vocabulary of words. Negative sampling simplifies this task by transforming the density estimation into a binary classification problem, which discriminates between genuine and spurious samples~\cite{Gutmann2010Noise}. Subsequently, Mikolov et al. presented a simplified version of NCE~\cite{Mikolov2013DistRepresent} that employs negative sampling to enhance the training of word2vec.  

Language models have become prevalent tools for estimating the probability distribution of the subsequent word within a given vocabulary, leveraging both the context and model parameters denoted as $\theta$.
The \textit{softmax} function plays a crucial role in determining the conditional distribution for the predicted word, which is denoted as $P_\theta^c(w)$, i.e.,
\begin{equation*}
P_\theta^c(w) = \frac{\exp(s_\theta(w,c))}{\sum_{w' \in V} \exp(s_\theta(\bar{w},c))} = \frac{u_\theta(w,c)}{Z_\theta^c}
\end{equation*}
where $s_\theta(w,c)$ is the score of a word $w$ in context $c$, $u_\theta(w,c) = \exp(s_\theta(w,c))$, and $Z_\theta^c$ is the partition function that is utilized to normalize the distribution to obtain a probability distribution over the vocabulary $V$, given context $c$. 
In contrast, Noise Contrastive Estimation avoids calculating the partition function under two assumptions~\cite{Mnih2012NCE}. First, NCE approximates the partition function $Z_\theta^c$ by introducing a parameter $z_c$ estimated from data for every empirical context $c$. Second, for networks with many parameters, setting $z_c = 1$ for all contexts is shown to be effective. In two-class training, NCE samples training data from the data set distribution $P_d^c(w)$ (denoted by $D = 1$) and $k$ noise samples from the noise distribution $P_n(w)$ (denoted by $D = 0$), which can be context-dependent or independent, as follows.

\begin{equation*}
\begin{aligned}
    p^c(w)& = P^c(D=1,w)+P^c(D=0,w) \\
    & = \frac{1}{k+1}[P^c_d(w)+kP_n(w)],
\end{aligned}
\end{equation*}
where;
\begin{equation*}
\begin{split}
    P^c(D=1,w)=\frac{1}{k+1}P^c_d(w), \\
    P^c(D=0,w)=\frac{k}{k+1}P_n(w).
\end{split}
\end{equation*}

Referring to NCE assumptions, the conditional likelihood of being a true distribution sample or noise sample is:
\begin{equation*}
\begin{split}
    p^c(D=1\mid w, \theta)=\frac{P^c(D=1,w)}{p^c(w)}=\frac{P^c_d(w)}{P^c_d(w)+kP_n(w)} ,
\end{split}
\end{equation*}
\begin{equation*}
\begin{split}
    p^c(D=0 \mid w, \theta)=\frac{P^c(D=0,w)}{p^c(w)}=\frac{kP_n(w)}{P^c_d(w)+kP_n(w)}.
\end{split}
\end{equation*}
For a model containing parameters $\theta$ such that given context $c$ its predicted probability from \textit{softmax} $P_\theta^c(w)$ approximates $P_d^c(w)$ in the dataset,  the conditional likelihood in terms of $\theta$ is:
\begin{equation*}
\begin{split}
    p^c(D = 1 \mid w, \theta) = \frac{P_\theta^c(w)}{P_\theta^c(w) + k P_n(w)}, \\  
    p^c(D = 0 \mid w, \theta) = \frac{k P_n(w)}{P_\theta^c(w) + k P_n(w)}.
\end{split}
\end{equation*}

The training of the model can be performed by maximizing the conditional log-likelihood based on the distribution of the dataset. In this context, $d$ represents the source distribution of the target word $w$, with $k$ negative samples. Consequently, the task can be formulated as a binary classification problem, characterized by the parameters $\theta$:
\begin{equation}
\begin{split}
    J^c(\theta) = \mathbb{E}_{{w}\sim P_d^c(w)} log\;p^c(D = 1 \mid w, \theta) + \\
    k \; \mathbb{E}_{{w}\sim P_n(w)} log \; p^c(D = 0 \mid w, \theta).
    \label{eqn:objective_funnction}
\end{split}
\end{equation}

The process of training knowledge graph embedding models through negative sampling involves treating nodes as words and their neighbors as context, akin to the approach used in word embedding for language modeling. 
Instead of explicitly modeling the conditional probability of all nodes, KGE models focus on learning knowledge representation by differentiating between positive and negative instances.
The quality of negative samples plays a significant role in the training process and subsequent performance of downstream tasks related to knowledge representation. Generating high-quality negative samples can enhance semantic learning and improve the efficiency of the training procedure.
Therefore, producing quality negatives is essential in improving the learning of semantics and smoothing the training process.

\subsection{The General Framework of KGRL}

\begin{figure}[]
\centering
    \includegraphics[width=0.5 \textwidth]{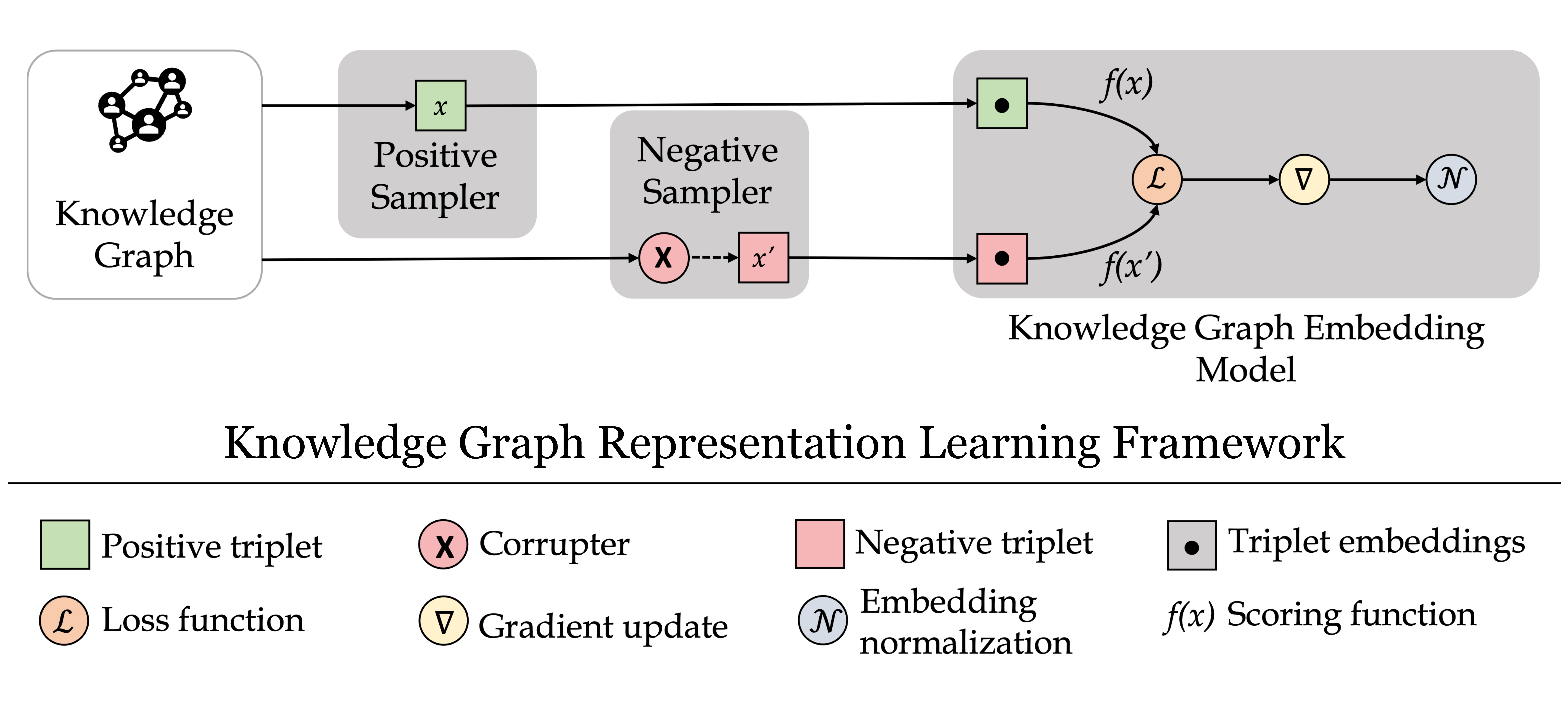}
\caption {The Knowledge Graph Representation Learning Framework aims to train a knowledge graph embedding model by optimizing the scoring function $f(x)$. The objective is to maximize the scores of positive samples $x=(h, r, t)$ while minimizing the scores of negative samples $x'$. This iterative process facilitates the acquisition of meaningful embeddings that capture the semantic relationships between entities and relations in the knowledge graph. } 
\label{fig:kge_framework}
\end{figure}

Knowledge graph representation learning is accomplished by utilizing the NCE framework, a comprehensive approach employed by numerous graph representation learning techniques. As depicted in Figure~\ref{fig:kge_framework}, this framework comprises three components: a KGE model, which is a trainable encoder generating embeddings based on a scoring function; a positive sampler, responsible for selecting positive samples; and a negative sampler, which generate negative samples for a given positive samples. Together, these components enable the generation of high-quality embeddings for use in KGRL optimizing the scoring function that maximizes observed triplets (positives) and minimizes non-observed triplets (negatives) simultaneously. 

During the training process of knowledge representation learning models, the closed-world assumption (CWA) and open-world assumption (OWA)~\cite{Reiter1978OWS} are taken into account. 
The CWA operates under the premise that unobserved facts are considered false. In contrast, the OWA adopts a more flexible stance by acknowledging that unobserved facts can either be missing or false. 
The OWA holds an advantage over the CWA, particularly given the inherent incompleteness of knowledge graphs. To optimize the KGE model, mini-batch optimization techniques and Stochastic Gradient Descent (SGD) are employed, aiming to minimize a specific loss function.

The general framework of knowledge representation learning is presented in Algorithm~\ref{alg:kg_embedding}. 
\begin{algorithm}[t]
	\DontPrintSemicolon
	  
	\KwInput{Triplet set $\mathcal{S} = \{(h, r, t)\}$, score function $f$ with parameters $\theta_{E}$, embedding dimension $d$, mini-batch size $m$, and training epochs ${T}$;}
	
	\KwOutput{Embedding model with parameters $\theta_{E}$}

        Initialize the embeddings for each $e \in \mathcal{E}$ and $r \in \mathcal{R}$.
	     
        \For{$ i = 1, ... , T$}
        {
            Sample a mini-batch $\mathcal{S}_{batch} \in \mathcal{S}$ of size $m$;
    
                \ForEach{$(h,r,t)$ $\in$ $\mathcal{S}_{batch}$ }
                {
                    Sample a negative triplet $(\bar h, r, \bar t) \in  \mathcal{\Bar{S}}_{(h,r,t)}$;

                    Update knowledge graph embeddings discriminating $(h,r,t)$ against ($\bar{h}, r, \bar{t}$) w.r.t the gradients using learning objective $L(\mathcal{E}, \mathcal{R})$.
                }
	}
	\caption{General framework of KGRL.}
	\label{alg:kg_embedding}
\end{algorithm}

The initialization of entity and relation embeddings is performed in step 1 of the training process. Subsequently, at each training epoch $i$, a mini-batch $\mathcal{S}_{\text{batch}}$ of size $m$ is drawn from the training set $\mathcal{S}$ during step 3. 
In step 5, a negative triplet $(\bar{h}, r, \bar{t})$ is selected from the set of manually created negative samples $\Bar{\mathcal{S}}(h, r, t)$. Here, $\Bar{\mathcal{S}}(h, r, t)$ is defined as the union of two sets: $\{(\bar{h}, r, t) \in \mathcal{S} \mid \bar{h} \in \mathcal{E}\}$ and $\{(h, r, \bar{t}) \in \mathcal{S} \mid \bar{t} \in \mathcal{E}\}$.
The embedding parameters are updated in step 6, guided by the learning objective $L(\mathcal{E}, \mathcal{R})$. This learning objective varies depending on the specific type of KGE approaches discussed in Section~\ref{subsection:kgrl}. For the typical geometric models, the objective function is defined as $L(\mathcal{E}, \mathcal{R}) = \sum_{(h,r,t) \in \mathcal{S}} [\gamma - f(h,r,t) + f(\bar{h},r,\bar{t})]{+}$. On the other hand, for the traditional tensor decomposition models, the objective function takes the form of $L(\mathcal{E}, \mathcal{R}) = \sum_{(h,r,t) \in \mathcal{S}} [l(+1, f(h,r,t)) + l(-1, f(\bar{h},r,\bar{t}))]$, where $f(\bar{h},r,\bar{t}) \notin \mathcal{S}$ represents a negative sample for $(h,r,t)$, and $l(\alpha, \beta) = \log(1 + \exp(-\alpha \beta))$ denotes the logistic loss function.
The process of generating and selecting negative samples poses a critical challenge in the optimization process. Negative sampling involves the selection of negatives from the complementary set $\Bar{\mathcal{S}}_{(h,r,t)}$.

\subsection{Importance of Negative Sampling in KGRL}
Based on the objective function of Noise Contrastive Estimation presented in Equation \ref{eqn:objective_funnction}, as well as prior research~\cite{Yang2020underneg}, the objective function for achieving optimal embeddings in KGRL can be expressed as follows:
\begin{dmath}
    J = \mathbb{E}_{{(u,v)}\sim P_d} log\; \sigma(\textbf{u}^\top \textbf{v}) + \mathbb{E}_{{v}\sim P_d(v)} [k \; \mathbb{E}_{{\bar{u}}\sim P_n(\bar{u}\mid v)} log \; \sigma(\bar{\textbf{-u}}^\top \textbf{v})]
    \\
    = \mathbb{E}_{{v}\sim P_d(v)} [\mathbb{E}_{{u}\sim P_d(u \mid v)} log\; \sigma(\textbf{u}^\top \textbf{v}) +  k \; \mathbb{E}_{{\bar{u}}\sim P_n(\bar{u} \mid v)} log \; \sigma(\bar{\textbf{-u}}^\top \textbf{v})],
\end{dmath}
where $P_d(u \mid v)$ is data distribution, $P_n(\bar{u}\mid v)$ is negative distribution, $\textbf{u}, \textbf{v}$ are embeddings for node $u$ and $v$ and $\sigma(.)$ is the \textit{sigmoid} function. 
Concerning the mathematical proof and the theorem of \textit{Optimal Embedding} presented in the work by Yang et al., it is demonstrated that maximizing the objective function $J$ can lead to optimal embedding for each pair of nodes $(u,v)$~\cite{Yang2020underneg}. 
This is equivalent to minimizing the objective function $J^{(v)}$ for each $v$, which can be achieved using the following approach.
\begin{align*}
J^{(v)} &= \mathbb{E}_{{u}\sim P_d(u \mid v)} log\; \sigma(\textbf{u}^\top \textbf{v}) +  k\mathbb{E}_{{\bar{u}}\sim P_n(\bar{u} \mid v)} log \; \sigma(\bar{\textbf{-u}}^\top \textbf{v})\\
&= \sum_{u} P_d(u \mid v) log\; \sigma(\textbf{u}^\top \textbf{v}) +  k \sum_{\bar{u}}P_n(\bar{u} \mid v) log \; \sigma(\bar{\textbf{-u}}^\top \textbf{v})\\
&= \sum_{u} [P_d(u \mid v) log\; \sigma(\textbf{u}^\top \textbf{v}) +  k P_n(u \mid v) log \; \sigma(\textbf{-u}^\top \textbf{v})]\\
&= \sum_{u} [P_d(u \mid v) log\; \sigma(\textbf{u}^\top \textbf{v}) +  k P_n(u \mid v) log(1 -\; \sigma(\textbf{u}^\top \textbf{v}))].
\end{align*}
For each $(u,v)$ pair, Let us consider the two Bernoulli distributions $P_{u,v}(x)$ and $Q_{u,v}(x)$ where;
\begin{align*}
    P_{u,v}(x = 1) &= \frac{P_d(u \mid v)}{P_d(u \mid v) + k P_n(u \mid v)}, \; \\
    Q_{u,v}(x = 1) &= \sigma(\textbf{u}^\top \textbf{v}).
\end{align*}
Then simplify the $J^{(v)}$ as follows:
\begin{equation*}
    J^{(v)} = - \sum_{u} (P_d(u \mid v) +  k P_n(u \mid v)) H(P_{u,v},Q_{u,v}),
\end{equation*}
where $H(p,q)= - \sum_{x \in X}p(x)log \; q(x)$ is the cross entropy between two distributions. According to the \textit{Gibbs Inequality}, the minimization of $J^{(v)}$ satisfies when $P=Q$ for each $(u,v)$ pair, which gives;
\begin{align*}
\frac{1}{1+e^{-\textbf{u}\top \textbf{v}}} &= \frac{P_d(u \mid v)}{P_d(u \mid v) + k \;P_n(u \mid v)}
\\
\textbf{u}\top \textbf{v} &= - log \frac{k.P_n(u \mid v)}{P_d(u \mid v)}.
\end{align*}

This demonstrates that both the positive ($P_d$) and negative distributions ($P_n$) have a similar impact on optimization, highlighting the importance of investigating negative sampling methods for achieving optimal embeddings in knowledge representation learning.

In addition to mathematical proof, the importance of negative sampling in KGRL can be examined from three main perspectives: training efficiency, managing class imbalance, and improving embedding quality as follows.
\begin{itemize}
    \item \textit{Efficient Training via Negative Sampling}:
    Negative sampling significantly improves the computational efficiency of KGE models. Given the immense size of knowledge graphs, evaluating all possible incorrect triples is impractical. Instead, negative sampling selects a manageable subset of negative samples, transforming the training process into a binary classification task. This allows the model to focus on informative negative examples, speeding up training and optimizing computational resources. As a result, training becomes more scalable, and the model converges faster without being overwhelmed by the size of the graph.
    \item \textit{Balancing Class Imbalance with Synthetic Negatives}:
    Managing class imbalance is essential in KGRL, as training datasets primarily consist of positive triples. The absence of explicit negative examples hinders the model's ability to differentiate valid triples from invalid ones. Negative sampling addresses this issue by generating synthetic negative triples, creating a more balanced dataset. This balance between positives and negatives improves the model’s generalization, reducing the risk of overfitting to positive samples and enhancing its ability to reject incorrect triples during evaluation.
    \item \textit{Improving Embedding Quality with Hard Negatives}:
    Embedding quality is enhanced when the model trains on hard negatives, i.e., negative triples that closely resemble positive triples in the embedding space. These challenging examples force the model to learn more refined and precise embeddings, capturing subtle differences between entities and relations. By focusing on hard negatives, the model learns to identify nuanced patterns in the knowledge graph, leading to more accurate embeddings and improved performance in tasks such as link prediction and entity classification.   
\end{itemize}

{\renewcommand{\arraystretch}{1.5}%
\begin{table*}[ht]
\caption{Categorization of negative sampling approaches for KGRL.}
\begin{center}
\begin{tabular}{lll}
\Xhline{1.3pt} 
\textbf{Category}           & \textbf{Subcategory} & \textbf{Strategies}  \\ \Xhline{1pt} 
\multirow{4}{*}{Static NS}  & Random               
                                & Uniform~\cite{Bordes2013Translating},
                                Random Corrupt~\cite{Peng2018RandomCorrupt}, 
                                Batch NS~\cite{Lerer2019BatchNS}
                            \\ \cdashline{2-3} 
                            & Probabilistic        
                                & \begin{tabular}[c]{@{}l@{}} Bernoulli~\cite{Wang2014Knowledge}, PNS~\cite{Kanojia2017pns}, FRR~\cite{Zhang2019MYX},  
                                SparseNSG~\cite{Cao2021SparseNSG}, Domain NS~\cite{xie2017DomainNS}, \\
                                ERDNS~\cite{Yao2023ERDNS},
                                Gibbs NS~\cite{Lijuan2024GNS}
                                \end{tabular}
                            \\ \cdashline{2-3} 
                            & Model-guided       
                                & 
                                NN~\cite{kotnis2017analysis},  NMiss~\cite{kotnis2017analysis},
                                LEMON~\cite{Alam2022Lemon},
                                LTS~\cite{Tian2022LTS}
                            \\ \cdashline{2-3} 
                            & Knowledge-constrained       
                                & \begin{tabular}[c]{@{}l@{}}
                                Type-Constraints~\cite{Krompas2015TypeNS}, STC~\cite{Xie2016STC}, RCWC~\cite{Wang2022KGBoost},
                                Trans$X_c$~\cite{Juan2019Transx}, 
                                Conditional-Constraints~\cite{Weyns2020ConditionalCF},\\
                                SANS~\cite{Ahrabian2020StructureAN},
                                Adaptive NS~\cite{Zhu2023Adaptive}, CDNS~\cite{Wang2019CDNS},
                                Concept-Driven~\cite{Yuan2023ConceptDriven},
                                GNS~\cite{lopez2023GNS}
                                \end{tabular}
                            \\ \hline
\multirow{3}{*}{Dynamic NS} & Model-guided       
                                & ANS~\cite{Qin2019ANS}, EANS~\cite{Je2022EntityAN}, 
                                HTENS~\cite{lin2023HTENS},
                                Type-augmented~\cite{Peng2023TypeAugment}
                            \\ \cdashline{2-3} 
                            & Self-adaptive     
                                & \begin{tabular}[c]{@{}l@{}} $\epsilon$-Truncated UNS~\cite{Sun2018Bootstrapping},
                                Truncated NS~\cite{Truncated2023Li} , ADNS~\cite{Alam2020AffinityDN},
                                DNS~\cite{Dash2019DNS},
                                HNS-SF~\cite{Che2022MixKGMF}, \\
                                HNS-CES~\cite{Che2022MixKGMF},
                                SNS~\cite{Islam2022SNS},
                                DSS~\cite{Nie2023DSSS}
                                \end{tabular}
                            \\ \cdashline{2-3} 
                            & Knowledge-constrained       
                                & Reason-KGE~\cite{Jain2021ReasonKGE}
                            \\ \hline
\multirow{2}{*}{Adversarial NS} & Standard             
                                & \begin{tabular}[c]{@{}l@{}}
                                KBGAN~\cite{Cai2018KBGANAL}, IGAN~\cite{Wang2018IncorporatingGF}, GraphGAN~\cite{Wang2018GraphGAN}, 
                                KCGAN~\cite{Zia2021KCGAN}, 
                                GN+DN~\cite{LIU2022GNDN}, 
                                \end{tabular}
                            \\ \cdashline{2-3}
                            & Knowledge-constrained              
                                & \begin{tabular}[c]{@{}l@{}}
                                KSGAN~\cite{Kairong2019KSGAN}, NKSGAN~\cite{Hai2020NKSGAN},
                                RUGA~\cite{Caifang2021RUGA},
                                NoiGAN~\cite{Cheng2020NoiGAN},
                                NAGAN-ConvKB~\cite{Le2021NAGAN-ConvKB}
                                \end{tabular}
                            \\ \hline
\multirow{3}{*}{Self Adversarial NS} & Direct 
                                & \begin{tabular}[c]{@{}l@{}} 
                                CANS~\cite{Shan2018ConfidenceAware},
                                Self-Adv.~\cite{Sun2019RotatEKG}, NSCaching~\cite{Zhang2019NSCachingSA}, LAS~\cite{Lei2019LAS}, 
                                    ASA~\cite{qin2021ASA},  ESNS~\cite{Yao2022ESNS} , 
                                \\
                                ABNS~\cite{si2023attention},
                                TANS~\cite{Feng2024TANS}             \end{tabular}                      
                            \\ \cdashline{2-3} 
                            & Model-guided      
                                & \begin{tabular}[c]{@{}l@{}} 
                                Self-Adv. SANS~\cite{Ahrabian2020StructureAN}, 
                                Self-Adv. EANS~\cite{Je2022EntityAN},
                                MCNS~\cite{Yang2020underneg}, 
                                MDNCaching~\cite{tiroshan2022mdncaching}, \\ TuckerDNCaching~\cite{tiroshan2022tuckerdncaching},
                                CCS~\cite{Han2023CCS}, 
                                HaSa~\cite{Zhang2024HaSa}
                                
                                \end{tabular}
                            \\ \cdashline{2-3} 
                            & Knowledge-constrained       
                                & Local-cognitive~\cite{huang2020rate}, CAKE~\cite{Niu2022cake}
                            \\ \hline
\multirow{2}{*}{Mix-up NS}  & Standard             
                                & \begin{tabular}[c]{@{}l@{}}
                                MixKG~\cite{Che2022MixKGMF}, DeMix~\cite{Chen2023DeMix},
                                M²ixKG~\cite{Che2024M2ixKG}
                                \end{tabular}
                            \\ \cdashline{2-3} 
                            & Memory-based
                                & DCNS~\cite{Zheng2024DCNS}
                            \\ \hline
\multirow{1}{*}{Text-based NS}  & Generative          
                                & \begin{tabular}[c]{@{}l@{}}
                                GHN~\cite{Qiao2023GHN}
                                \end{tabular}
                            \\ \Xhline{1.3pt} 
\end{tabular}
\end{center}
\label{table:ns}
\end{table*}

\subsection{Dimensions of Negative Sampling}
The success of negative sampling in knowledge graph representation learning depends on several critical factors. These factors help evaluate the performance of different negative sampling methods, and we classify them into five key dimensions:

\begin{itemize}
    \item \textit{Efficiency}: 
    Efficiency refers to the computational cost and resource consumption of the negative sampling method. An efficient method should have low time complexity to ensure quick generation of negative samples and minimal memory overhead to handle large-scale knowledge graphs without bottlenecks in processing.
    \item \textit{Effectiveness}:  Effectiveness is the ability of the sampling method to generate negative samples that improve the training of knowledge graph embeddings. High effectiveness means that the negative samples drive the model toward learning better representations by challenging it to differentiate between positive and negative triples, avoiding trivial or overly easy samples that do not contribute to learning.
    \item \textit{Stability}: Stability measures how reliably the method performs across different datasets and settings. A stable negative sampling approach produces consistently good results, regardless of variations in dataset size, structure, or complexity. It should also be robust against the risk of overfitting or poor generalization to unseen data.
    \item \textit{Adaptability}: Adaptability reflects the method's flexibility to work with different types of knowledge graphs and models. A highly adaptable method requires minimal reliance on specific assumptions or additional external information (e.g., entity types, external resources), making it suitable for a wide range of applications and models in KGRL.
    \item \textit{Quality}: Quality refers to the relevance and informativeness of the negative samples. High-quality negative samples should be semantically meaningful and challenging, helping the model learn to distinguish between true and false triples. The negative samples should reflect the underlying structure of the knowledge graph and promote better generalization by capturing important variations in the data.
\end{itemize}

These dimensions highlight the key considerations when evaluating and selecting negative sampling methods for KGRL. While no single approach optimally satisfies all five dimensions, understanding the trade-offs between them can guide the choice of the most appropriate method for specific tasks or datasets.
%
%

\begin{figure*}[]
\centering
    \includegraphics[width=1.0\textwidth]{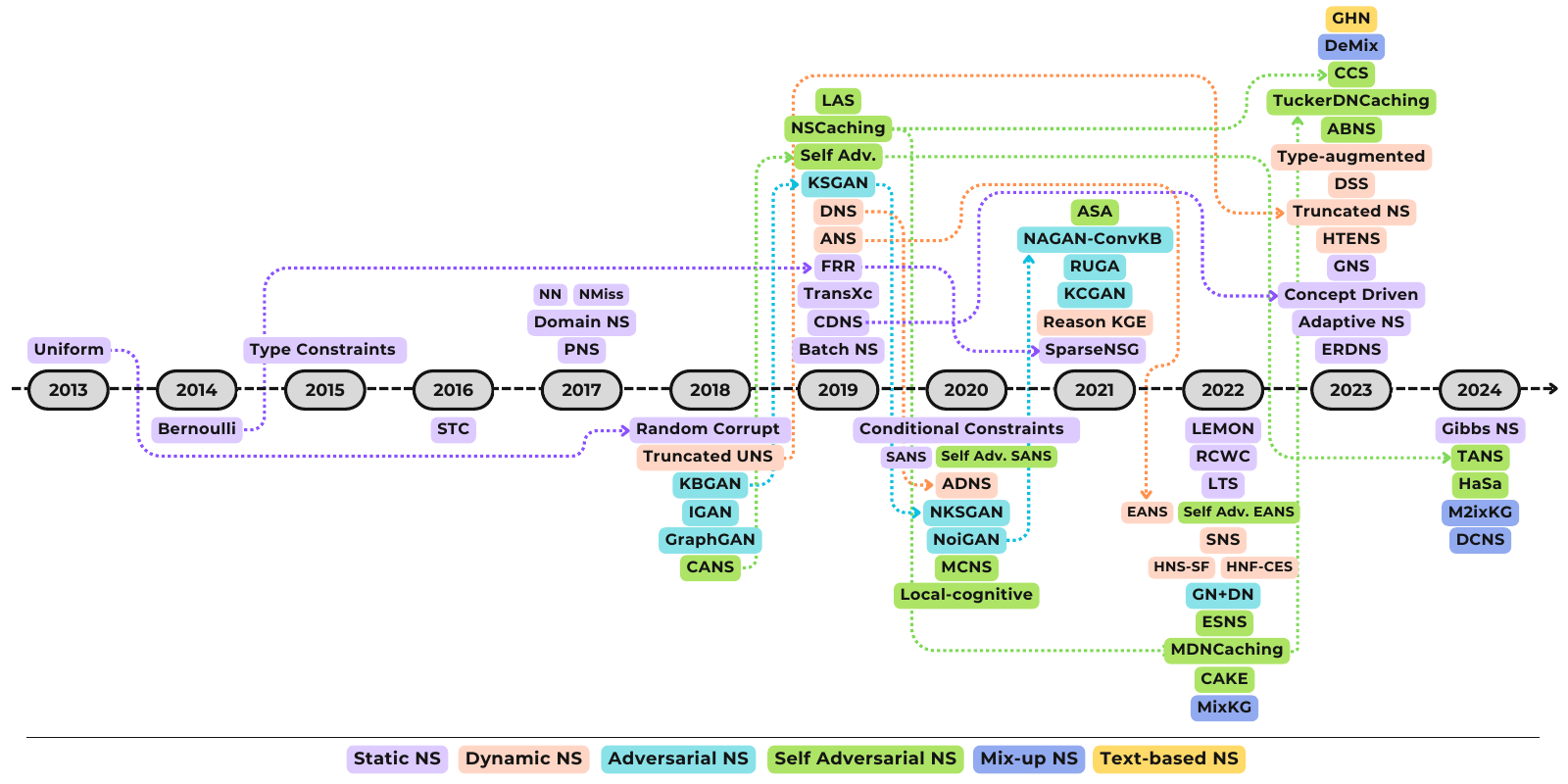}
\caption {Timeline of Negative Sample in Knowledge Graph Representation Learning. Dotted arrows indicate that the target method extends the source method. }
\label{fig:timeline}
\end{figure*}

\section{Negative Sampling in KGRL}
\label{section:negative_sampling}

This section systematically analyzes negative sampling techniques used in knowledge graph representation learning. Negative sampling is essential for generating negative distributions and developing strategies to select negative samples from a pool of candidates. The complexity of these techniques ranges from simple methods like random sampling to more sophisticated approaches that account for factors such as model state, sample complexity, and dataset characteristics.

To offer a clearer comparison, we classify these methods into six primary categories, each defined by its approach to negative sample generation, as shown in Table~\ref{table:ns}. Below, we describe each category and its distinguishing features:

\begin{itemize}
    \item \textit{Static Negative Sampling (Static NS)}:
This method generates negative samples before training and keeps them constant throughout the process. Common approaches include uniform and frequency-based sampling, where negatives are selected randomly or based on their occurrence frequency. Static NS is computationally efficient and easy to implement, but it lacks in effectiveness, as the negative samples do not evolve with the embeddings, potentially limiting the model's ability to learn from challenging examples.

    \item \textit{Dynamic Negative Sampling (Dynamic NS)}:
    Unlike static NS, dynamic methods generate negative samples during training, adapting the selection as the model’s embeddings change. This ensures that the model encounters increasingly challenging negatives, leading to better learning and more accurate embeddings. Although dynamic NS is more computationally intensive, it offers superior performance, particularly in complex graphs, by keeping the negative samples aligned with the model’s evolving understanding.

    \item \textit{Adversarial Negative Sampling (Adversarial NS)}:
    Adversarial NS selects negatives that closely resemble positive triples, making it harder for the model to distinguish between them. This is achieved by using an adversarial model or loss function to generate negatives near positive triples in embedding space. By confronting the model with difficult samples, adversarial NS improves its ability to learn nuanced distinctions, though it requires additional computational resources due to the complexity of the adversarial process.

    \item \textit{Self-Adversarial Negative Sampling (Self-Adversarial NS)}:
    Self-adversarial NS simplifies the adversarial approach by allowing the model to determine which negatives are the hardest referring to embeddings of the model itself. The model dynamically assigns higher importance to challenging negatives, prioritizing them in the learning process. This technique balances efficiency and the generation of informative samples, without needing an external adversarial mechanism, enhancing the model's ability to detect complex patterns in the knowledge graph.
    
    \item \textit{Mix-up Negative Sampling (Mix-up NS)}:
    Mix-up NS combines data augmentation with negative sampling to improve training. This method creates synthetic negatives by interpolating between hard negative triplets, leading to more difficult samples for the model to learn from. Mix-up NS enriches the training dataset and helps the model capture complex relationships, though it introduces additional computational overhead compared to simpler techniques.
    
    \item \textit{Text-based Negative Sampling (Text-based NS)}:
    This approach leverages semantic information from entity and relation descriptions to generate negative samples that are contextually relevant and challenging. By incorporating textual embeddings, text-based NS ensures that the negatives are semantically close to positive examples, pushing the model to make finer distinctions. This method is especially beneficial for improving the quality of embeddings in domains where textual information complements the structured knowledge graph.
    
\end{itemize}

The timeline in Figure~\ref{fig:timeline} shows the evolution of negative sampling techniques in KGRL over the past decade. Early methods focused on static approaches but advances in dynamic, adversarial, mix-up, and text-based strategies reflect the increasing sophistication of negative sampling. These innovations continue to drive improvements in model performance, emphasizing the importance of balancing structural and semantic learning in knowledge representation.

\subsection{Static NS}

Static negative sampling is a widely used approach in KGRL to generate negative samples for training models to differentiate between true and false triples. This method operates under the closed-world assumption, where any triple absent from the knowledge graph is assumed to be false. However, applying the CWA introduces challenges due to the inherent incompleteness of knowledge graphs, potentially leading to false negatives. Additionally, static sampling can result in a large number of incorrect facts, making the set of negative instances considerable, with a complexity of $(O(N^2))$.

\begin{figure}[]
\centering
    \includegraphics[width=0.5\textwidth]{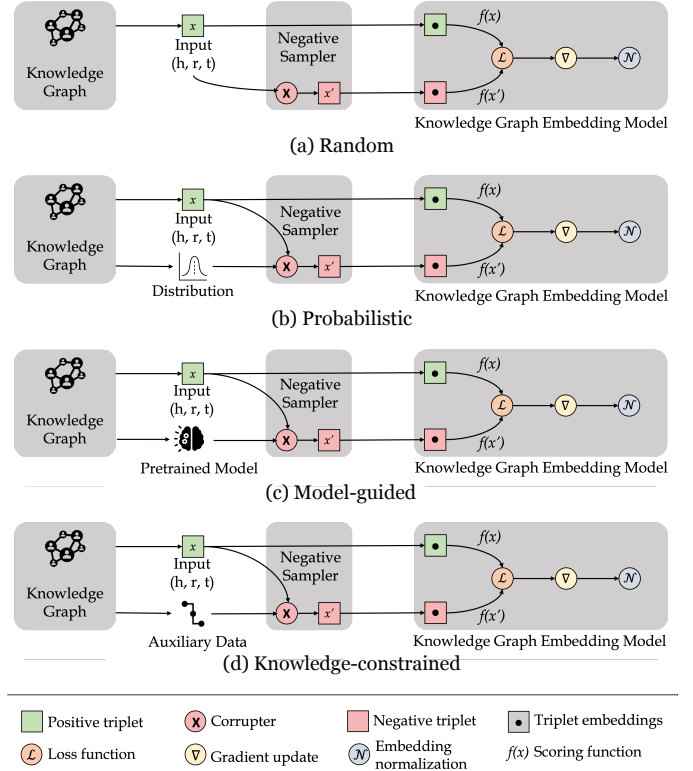}
\caption {Overview of subcategories of static negative sampling methods and steps of training a knowledge graph embedding model over a positive instance $x=(h, r, t)$ and a corrupted negative $x'$ from different static negative sampling methods.}
\label{fig:static_ns}
\end{figure}

Figure~\ref{fig:static_ns} presents an overview of various static negative sampling methods and the steps involved in training a KGE model. Below are the four main variants of static negative sampling commonly used in KGRL:
\begin{itemize}
    \item \textit{Random}: This straightforward approach generates negatives by randomly corrupting either the head or tail entity of positive triples. As illustrated in Figure~\ref{fig:static_ns}(a), this method is computationally efficient but often leads to trivial negatives that are easily distinguishable from positive triples, thus providing less informative samples for training.
    \item \textit{Probabilistic}:
    Instead of purely random selection, this variant introduces a fixed distribution $P$ to choose negatives based on criteria such as entity frequency or other statistical properties of the data. This method, shown in Figure~\ref{fig:static_ns}(b), tends to generate more representative and balanced negative samples, which can improve training by focusing on more meaningful negatives.
    \item \textit{Model-guided}: In this variant, a pre-trained model, i.e., embedding model or language model, is employed to generate a fixed set of quality negative samples, as seen in Figure~\ref{fig:static_ns}(c). This approach ensures that the negative samples are challenging and relevant, as they are selected based on the learned features of the knowledge graph or other supporting models. However, it introduces an additional computational overhead due to the necessity of pre-training the model before generating these negatives.
    \item \textit{Knowledge-constrained}: This approach leverages auxiliary data, such as entity type information, semantic relationships, or pre-generated structural data, to produce negative samples, as shown in Figure~\ref{fig:static_ns}(d). By incorporating more contextually relevant information, this method can improve the quality of the negative samples and lead to better model performance.
\end{itemize}

\subsubsection{Random} 
Uniform negative sampling~\cite{Bordes2013Translating} is one of the most widely used methods in KGRL. It generates negative samples by randomly replacing either the head $h$ or the tail $t$ entity of a positive triple $(h, r, t)$ with another entity from the entity set $\mathcal{E}$, using a uniform distribution. This approach operates under the Local Closed-World Assumption (LCWA), where any triple not present in the knowledge graph is assumed to be false. However, Uniform negative sampling has notable limitations, such as producing false negatives.  

Random Corrupt~\cite{Peng2018RandomCorrupt} builds upon Uniform sampling by not only corrupting the head or tail entity but also introducing the corruption of the relation $r \in \mathcal{R}$ to generate negative triples. This expansion aims to enhance the representation of relation information, thereby improving the diversity and quality of negative samples.

Batch NS~\cite{Lerer2019BatchNS} introduces a novel approach by reusing samples within the same mini-batch as negative candidates. This method extends the conventional uniform sampling technique by promoting sample reuse, effectively addressing the challenge of resource allocation in negative sampling. By leveraging this strategy, Batch NS provides a more efficient way of generating negative samples without increasing computational overhead.

\subsubsection{Probabilistic}
Bernoulli negative sampling~\cite{Wang2014Knowledge} improves KGRL performance by adjusting the probability of substituting the head or tail entity in a positive triple based on the relation's mapping properties. Specifically, the probability of replacing the head entity $P_h$ is computed as $P_h = tph / (tph + hpt)$, where $tph$ refers to the average number of tail entities per head entity, and $hpt$ is the average number of head entities per tail entity. The tail substitution probability $P_t$ is similarly calculated as $P_t = hpt / (tph + hpt)$. This approach allows for the generation of more contextually appropriate negatives based on relation-specific probabilities.

PNS (Probabilistic Negative Sampling)~\cite{Kanojia2017pns} offers a more adaptive solution in scenarios where data is sparse or imbalanced. PNS introduces a training bias parameter $\beta$, which accelerates the creation of corrupted triples by considering entity and relation distributions in the knowledge base. The parameter $\beta$ allows the model to adjust the probability of selecting potential negative instances, improving sampling efficiency in diverse datasets.

FRR (Flexible Relation Replacement)~\cite{Zhang2019MYX} extends the Bernoulli method by introducing a probability $\alpha$ for replacing relations in a triple. The relation replacement probability is defined as $\alpha = \mid \mathcal{R} \mid / (\mid \mathcal{R} \mid + \mid \mathcal{E} \mid)$. FRR also refines the entity replacement probabilities: $P_h = (1 - \alpha) \frac{tph}{tph + hpt}$ and $P_t = (1 - \alpha) \frac{hpt}{tph + hpt}$. This approach ensures a more balanced distribution of negative samples by incorporating relation-specific probabilities.

Building on FRR, SparseNSG (Sparse Negative Sample Generation)~\cite{Cao2021SparseNSG} enhances the relation replacement strategy by introducing an additional statistic, $C_{Ent}(r)$, which captures the number of entities associated with a relation $r$. This statistic adjusts the relation replacement probability to $\alpha = \frac{C_{Ent}(r)}{1 + C_{Ent}(r)} \times \frac{\mid \mathcal{R} \mid }{\mid \mathcal{R} \mid + \mid \mathcal{E} \mid}$. SparseNSG is particularly effective in large, sparsely connected knowledge graphs, where it improves model performance by considering the structure and connectivity of relations.

Conversely, Domain NS~\cite{xie2017DomainNS} introduces a domain-specific negative sampling method that leverages the characteristics of entities within specific domains. The relation-dependent probability $P_r$ selects entities from the same domain as the relation with probability $P_r$, and from the entire entity set $\mathcal{E}$ with probability $1 - P_r$. The probability $P_r$ is calculated as $P_r = \min(\frac{\lambda \mid M^H \mid \mid M^T \mid}{\mid N_r \mid}, 0.5)$, where $M^H$ and $M^T$ are the domains associated with the head and tail entities, respectively, and $N_r$ is the set of edges related to the relation $r$. The hyperparameter $\lambda$ controls the balance between domain specificity and generality in negative sampling.

ERDNS (Entity-Relation Distribution-Aware Negative Sampling)~\cite{Yao2023ERDNS} offers an adaptive approach by adjusting the number of negative samples for entity-relation (ER) pairs based on their frequency in the dataset. Frequently occurring ER pairs are assigned fewer negative samples, while rarer pairs receive more. This method mitigates the risk of generating false negatives by directly considering the distribution of ER pairs, unlike Bernoulli sampling, which focuses on average head-to-tail frequencies ($hpt$ and $tph$). ERDNS assigns distinct negative sample counts $N(e,r)$ to each ER pair, improving the quality of negative samples, particularly for infrequent entity-relation combinations.

Gibbs NS (Gibbs Negative Sampling)~\cite{Lijuan2024GNS} integrates Gibbs sampling with path-based strategies to generate high-quality negative samples. By analyzing the distributional characteristics of entities within the KG, Gibbs NS enhances the representation of relations. The method first selects entities using Gibbs sampling to corrupt triples and then partitions positive and negative samples based on head and tail entities. Two directional mapping matrices are constructed, and metric learning is applied to assess the similarity between these matrices. This sophisticated sampling approach improves the expressiveness of the model and its ability to capture complex relationships in the knowledge graph.

\subsubsection{Model-guided}
NN (Nearest Neighbor) negative sampling~\cite{kotnis2017analysis} selects negative samples that are close to the positive triple in the embedding space. This proximity forces the KGE model to distinguish between positive instances and semantically similar negative samples. On the other hand, NMiss (Near Miss) negative sampling~\cite{kotnis2017analysis} selects negative candidates ranked higher than positive triples, pushing the model to improve its discrimination capabilities by focusing on these harder negative samples.

LEMON (LanguagE MOdel for Negative Sampling)~\cite{Alam2022Lemon} enhances negative sampling by using pre-trained language models to create clusters of neighborhood entities based on textual information. The method incorporates Sentence-BERT embeddings\cite{Reimers2019SentenceBERTSE} and applies PCA~\cite{Hervé2010PCA} for dimensionality reduction, followed by the K-means++ clustering algorithm~\cite{David2007kmeans} to select challenging negative samples from similar entities.

LTS (LDA Topic Similarity)~\cite{Tian2022LTS} utilizes topic modeling, specifically Latent Dirichlet Allocation (LDA), to group entities based on their semantic topics. During negative sampling, entities from the same topic are used to corrupt positive triples, ensuring that the generated negatives are semantically similar to the positives.

\subsubsection{Knowledge-constrained}
Type-Constraints negative sampling~\cite{Krompas2015TypeNS} leverages RDF-Schema type constraints to generate negative samples that respect the class-based domain (\texttt{rdfs:domain}) and range (\texttt{rdfs:range}) restrictions defined in the KG. The Type-Constraints LCWA variant extends this by applying the constraints on an instance level, reducing inconsistencies in relation types.

STS (Soft Type Constraint)~\cite{Xie2016STC} improves negative sampling by increasing the likelihood of selecting negative entities that share the same type as the positive sample, guided by relation-specific type information. The approach balances diversity and similarity by adjusting the constraint's softness through a hyperparameter $k$ in the probability formula $P(\bar{e} \in \mathcal{E}) = ((k + 1)|\mathcal{E}_c|) / (|\mathcal{E}| + k|\mathcal{E}_c|)$, where $c$ is the golden type for the entity in the triple, $\mathcal{E}_c$ is the set of entities with type $c$.

RCWC (Range-Constrained with Co-occurrence)~\cite{Wang2022KGBoost} defines a co-occurrence measure between entities within the same relation to generate negative samples. By excluding corrupted entities that frequently co-occur with positive ones, RCWC minimizes the risk of producing false negatives and generates more informative samples.

Trans$X_c$~\cite{Juan2019Transx} incorporates domain and range constraints into conventional translation-based embedding models. It employs logistic regression classifiers to capture the specific characteristics of relation domains and ranges and generates negative samples by ensuring that the replaced head or tail entity maintains the domain or range compatibility of the original relation.

Conditional-Constraints negative sampling~\cite{Weyns2020ConditionalCF} extends schema-based approaches by incorporating OWL (Web Ontology Language) restrictions. This method uses domain and range constraints, as well as conditional constraints extracted from OWL restrictions, i.e., \texttt{owl:onProperty} that implies the relationship is part of an \texttt{owl:Restriction}, to generate high-quality negative samples.

SANS (Structure Aware Negative Sampling)~\cite{Ahrabian2020StructureAN} leverages the structural properties of the KG, selecting negative candidates based on the node's neighborhood structure. It examines the k-hop neighbors of a given node and utilizes the connections between nodes to generate informative negative samples that reflect the complex structure of the KG.

Adaptive NS~\cite{Zhu2023Adaptive} adapts the negative sampling process based on the frequency of positive samples, assigning more negative samples to low-frequency triples. By prioritizing less frequently observed triples, this method ensures that the model pays more attention to sparse samples, improving performance on long-tail relations.

CDNS (Concept Domain Negative Sampling)~\cite{Wang2019CDNS} refines the negative sampling process by integrating structured KG information, i.e. from encyclopedic KGs, with semantic concepts from lexical KGs. It samples negative heads or tails based on the shared concepts of the entities, leveraging prior probability distributions derived from their semantic similarity.

Concept-driven negative sampling~\cite{Yuan2023ConceptDriven} extends CDNS by enhancing commonsense understanding in knowledge graph completion tasks. It introduces concept constraints derived from lexical knowledge graphs (i.e., Probase~\cite{Wu2012Probase}) and generates negative samples using a hybrid strategy that balances concept-based and fact-based constraints. 

GNS (Good Negative Sampling)~\cite{lopez2023GNS} utilizes lightweight ontological axioms, such as relation domains, ranges, and functional properties, to guide the generation of high-quality negative triples. It also incorporates both object and type relations from the RDF model to improve the robustness of the sampling process.

\subsubsection{Pros and Cons.}
\textbf{Efficiency} is critical for avoiding computational bottlenecks during training. Random negative sampling stands out as the most efficient due to its simplicity and low overhead, allowing for rapid generation of samples even in large-scale knowledge graphs. Probabilistic negative sampling increases computational costs slightly due to probability adjustments, while Model-guided and Knowledge-constrained methods are the most computationally expensive as they require pre-trained models or auxiliary knowledge, which can slow down training, especially on large datasets. 
\textbf{Effectiveness} reflects how well the sampling method generates useful negatives that challenge the model. Random negative sampling often produces trivial negatives, such as mismatched entity types, leading to reduced effectiveness. In contrast, Probabilistic, Model-guided, and Knowledge-constrained methods offer more challenging negatives by leveraging probability distributions or external knowledge, making them more effective for training. 
\textbf{Stability} refers to how consistently these methods perform across datasets. Random negative sampling is less stable due to random entity selection, whereas Probabilistic negative sampling is more stable as it accounts for frequency distribution. However, Model-guided and Knowledge-constrained methods can be unstable if the external models or data sources are of poor quality or incomplete. 
\textbf{Adaptability} is a measure of the method's flexibility across different knowledge graphs and models. Random and Probabilistic negative sampling are highly adaptable as they do not rely on specific assumptions or external resources, whereas Model-guided and Knowledge-constrained methods are less adaptable, as they depend on the availability and relevance of external knowledge. 
Finally, \textbf{Quality} concerns the relevance of the generated negatives. Random negative sampling often suffers from poor quality, sometimes generating false negatives, whereas Probabilistic methods produce better-quality negatives by focusing on more challenging cases. Model-guided and Knowledge-constrained methods provide high-quality, semantically meaningful negatives but may still suffer from false negatives.

\subsection{Dynamic NS}

\begin{figure}[]
\centering
    \includegraphics[width=0.5\textwidth]{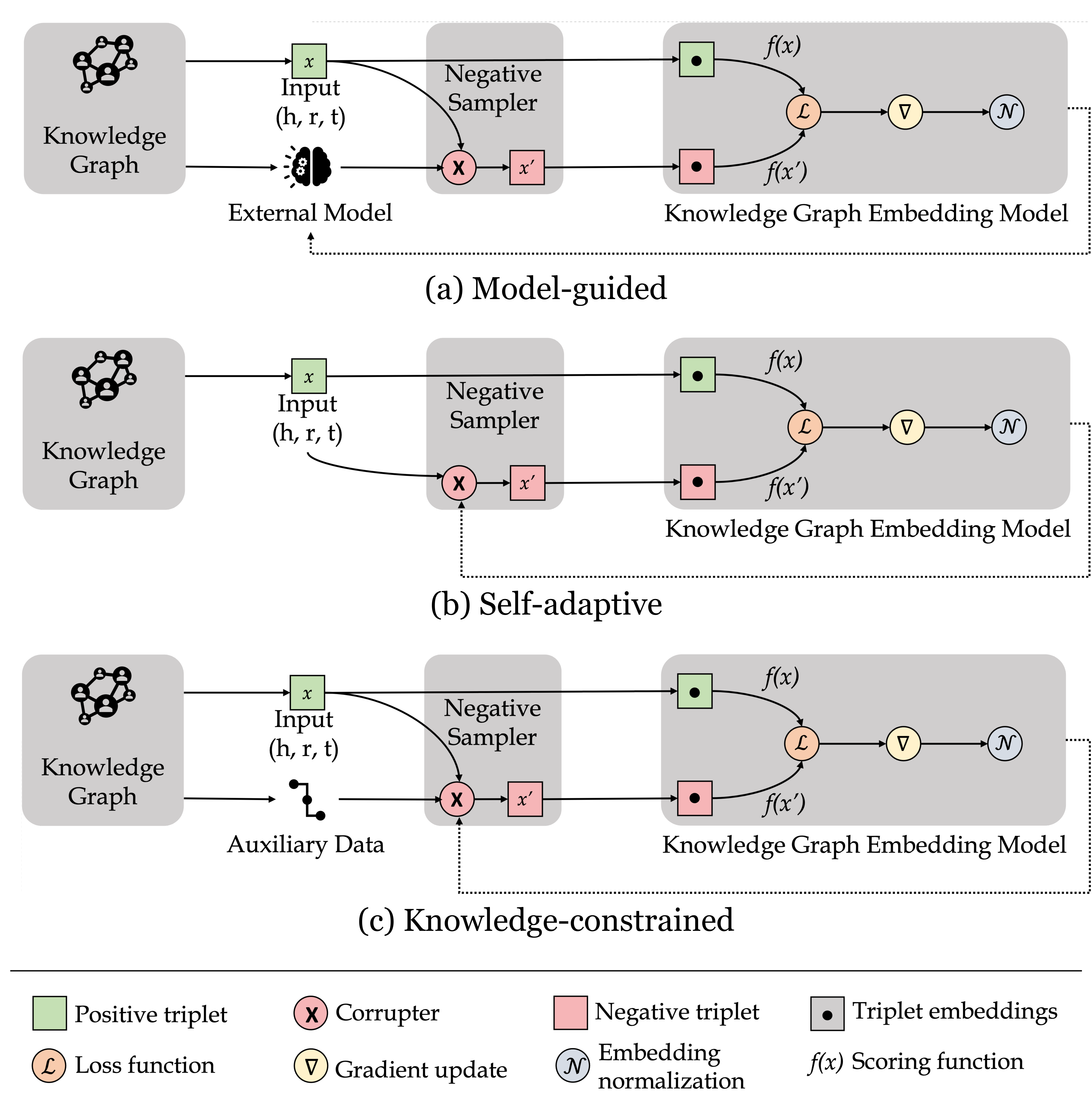}
\caption {Overview of subcategories of dynamic negative sampling methods and steps of training a knowledge graph embedding model over a positive instance $x=(h, r, t)$ and a corrupted negative $x'$ from different dynamic negative sampling methods.}
\label{fig:dynamic_ns}
\end{figure}

In KGRL, static negative sampling methods often generate trivial negative samples that can hinder the training process and lead to vanishing gradient problems. To overcome these limitations, dynamic negative sampling methods have been proposed, which adapt to the changing nature of the embedding space, providing more informative negative samples. These methods optimize sample selection by focusing on dynamic distributions within the negative sample space.

Figure~\ref{fig:dynamic_ns} provides an overview of the dynamic negative sampling approaches, categorizing them into three distinct subcategories below.

\begin{itemize}
    \item \textit{Model-guided}: This approach utilizes external machine learning models that continuously update to reflect the evolving characteristics of the target KGE model, as illustrated in Figure~\ref{fig:dynamic_ns}(a). These methods generate negative samples that adapt in conjunction with the model, thereby ensuring sustained quality and relevance throughout the training process.
    \item \textit{Self-adaptive}: 
    This method generates negative samples directly from the changing dynamics of the target embedding space, as depicted in Figure~\ref{fig:dynamic_ns}(b). By focusing solely on the internal alterations of the embedding space, this approach fosters a more responsive sampling mechanism that aligns with the model's progression.
    \item \textit{Knowledge-constrained}: 
    This variant leverages external data sources, such as schema information, type constraints, and ontology-driven methods, to enhance the quality of negative samples. As shown in Figure~\ref{fig:dynamic_ns}(c), these techniques integrate supplementary knowledge to dynamically refine negative samples, utilizing insights from both the knowledge graph and external datasets to improve sampling efficacy.
\end{itemize}
\subsubsection{Model-guided}
ANS (Adaptive Negative Sampling)~\cite{Qin2019ANS} generates negative samples using the K-means clustering model~\cite{Hartigan1979KMeans}. Initially, ANS groups entities into clusters based on similarity. For each positive triple, a negative entity is selected from the corresponding cluster of the head or tail entity. A lazy update mechanism refreshes the clustering model after several epochs of training, accommodating changes in the knowledge graph embeddings.

EANS (Entity-Aware Negative Sampling)~\cite{Je2022EntityAN} enhances ANS by utilizing a Gaussian distribution in the aligned entity index space to sample negative entities resembling positive ones. EANS groups entities with K-means and maintains an index mapping to avoid recalculating distances. It replaces the uniform distribution for corrupting positive triplets with a Gaussian distribution, also employing lazy updates for cluster embeddings.

HTENS (Hierarchical Type Enhanced Negative Sampling)~\cite{lin2023HTENS} incorporates hierarchical type information and co-occurrence patterns to produce high-quality negative samples. It utilizes a \textit{Hierarchical Type Similarity Score}, i.e., entity similarity based on hierarchical type information, considering sub-types at different granularities, and a \textit{Cooccurrence Similarity Score}, i.e., similarity through entity-relation co-occurrence patterns, combining them through a neural network sampler to calculate final sampling probabilities. The sampler is trained using Kullback-Leibler divergence loss to align with quality estimates from the KGE model.

Type-augmented negative sampling~\cite{Peng2023TypeAugment} introduces a type-constrained negative sampling approach without explicit type information. It constructs candidate sets of homogeneous and non-homogeneous entities, using the underlying model's embedding function to map elements into two vector spaces for relation-entity and relation-type triples. A type compatibility function based on semantic similarity constrains type features, and a relation-specific hyperplane projection distinguishes multiple entity types. The method dynamically selects corrupting entities, updating proportions as the model evolves, addressing the limited availability of homogeneous entities in sparse graphs.

\subsubsection{Self-adaptive}
In contrast to uniform negative sampling, $\epsilon$-Truncated UNS ($\epsilon$-Truncated Uniform Negative Sampling)~\cite{Sun2018Bootstrapping} constrains sampling to a specific set of candidates by selecting the $s$-nearest neighbors in the embedding space, where $s = \lceil (1 - \epsilon)N \rceil$ and $\epsilon \in [0, 1)$. This method uses \textit{cosine similarity} among embeddings to identify similar neighbors for the candidate set.

Truncated NS (Truncated Negative Sampling)~\cite{Truncated2023Li} integrates $\epsilon$-Truncated UNS while expanding the similarity measurement to include both structural and attribute embedding vectors of an entity $e$. The structural embedding ($\textbf{e}_{s}$) captures underlying structural characteristics, while the attribute embedding ($\textbf{e}_{a}$) reveals latent semantic information. The similarity measure is defined as $SIM = \sum_{e \in G_1 \bigcup G_2} [1 - sim(\textbf{e}_{s}, \textbf{e}_{a})]$, where $sim(.,.)$ denotes \textit{cosine similarity}, and $G_1$ and $G_2$ refer to the two knowledge graphs involved in entity alignment.

DNS (Distributional Negative Sampling)~\cite{Dash2019DNS} capitalizes on the observation that entities of the same type in a knowledge graph often share multiple relations. This method replaces entities in a given triple with others of the same type, generating meaningful assertions. DNS utilizes learned entity embeddings from the target KGE model to capture distributional properties, employing \textit{cosine similarity} among embeddings to approximate semantic similarity among candidate entities.

ADNS (Affinity Dependent Negative Sampling)~\cite{Alam2020AffinityDN} generates negative triples based on entity affinities, extending DNS principles. It calculates \textit{cosine similarity} between the candidate entity and all others, creating an affinity vector used as a probability vector for selection. The affinity vector is updated at each epoch to reflect changes in KG embeddings.

HNS-SF (Hard Negative Samples-Score Function) and HNS-CES (Hard Negative Samples-Correct Entity Similarity) methods were introduced for hard negative sampling in MixKG~\cite{Che2022MixKGMF}. HNS-SF computes scores for candidate negatives using a scoring function and selects the $k$ negative triplets with the highest scores. HNS-CES calculates similarity scores through the dot product ($s(\bar{t}) = t \cdot \bar{t}$) and selects the $k$ candidates with the highest scores as negative triplets.

SNS (Simple Negative Sampling)~\cite{Islam2022SNS} achieves a balance between exploration and exploitation by uniformly selecting negative candidates while excluding known positives. To mitigate over-exploitation, SNS introduces a data structure called the least recently selected (LRS), denoted as $LRS[q'_h,q'_t]$. When considering negative candidates for the tail entity, $q'_1(t)$ is modified to include LRS samples: $q'_1(t) := q'_1(t) \cup LRS[q'_t]$. Each negative sample's probability is determined by the distance score, calculated as the Euclidean distance between the corrupted tail entity $t$ and new entities $\bar{t}_i$, denoted as $d(t, \bar{t}_i) = \left\| {\textbf{t} - \bar{\textbf{t}}_i} \right\|$. The sampling probability score for each candidate negative $(h, r, \bar{t}_i) \in q'_1(t)$ is defined as $P(h,r,\bar{t}_i) = \frac{exp(1 / d(t, \bar{t}i))}{\sum_{j=1}^{N} exp(1 / d(t, t_j'))},$ and SNS ranks candidates based on $P(h,r,\bar{t}_i)$ for updating $LRS[q'_h, q'_t]$.

DSS (Dynamic Semantic Similarity)~\cite{Nie2023DSSS} presents a negative sampling method that prioritizes negative triples with high semantic similarity to positive triples. This approach consists of selecting the entity to replace based on distinct probabilities for head and tail replacements, thereby reducing low-quality negatives. Semantic similarity is measured via \textit{cosine similarity}, and only the top-$m$ most semantically similar negative triples are retained for training. Each epoch dynamically updates the candidate set to ensure the continued availability of high-quality negative samples.

\subsubsection{Knowledge-constrained}
ReasonKGE~\cite{Jain2021ReasonKGE} aims to enhance negative sampling in KGE models through an ontology-driven approach. This method utilizes symbolic reasoning techniques to identify and capture inconsistencies in predictions made by KGE models, which are then employed as negative samples in subsequent training iterations. Initially, any available negative sampling procedure trains the KGE model, followed by the selection of predictions leading to inconsistencies when added to the KG. By targeting these inconsistencies, ReasonKGE effectively identifies weaknesses in the KGE model and generates negative samples iteratively, improving subsequent training iterations.
\subsubsection{Pros and Cons.}
In terms of \textbf{Efficiency}, dynamic methods face challenges due to the need to continuously evaluate the evolving target embedding space, which increases computational costs. Self-adaptive sampling is the most efficient, as it leverages current embeddings directly, avoiding external model training or processing additional data. In contrast, Model-guided methods incur higher costs due to model updates, and Knowledge-constrained methods require extra resources to integrate external knowledge, making them less efficient for large-scale graphs.
Regarding \textbf{Effectiveness} dynamic methods improve training by generating more meaningful negative samples. Model-guided sampling is highly effective, producing semantically rich samples by adjusting to the evolving model. Knowledge-constrained methods also perform well, using external knowledge like type constraints to create challenging samples. Self-adaptive sampling may lack the semantic richness of the other methods but still outperforms static sampling by evolving with the model.
For \textbf{Stability} dynamic methods tend to be less stable due to continuous embedding updates. Self-adaptive sampling is more stable, relying solely on the target model's embeddings, although rapid changes can still lead to false negatives. Model-guided methods are less stable, as they depend on the accuracy of external models, which can fluctuate. Knowledge-constrained methods show moderate stability, as they rely on relatively static external information, though inconsistencies in the data can cause variability.
In terms of \textbf{Adaptability}, self-adaptive sampling is the most flexible, requiring no external models or data, making it applicable to various knowledge graphs and models. Model-guided methods are less adaptable due to their dependence on external models, limiting their use in domains where such models are unavailable. Knowledge-constrained methods fall in between, as their reliance on external knowledge can hinder adaptability if the information is domain-specific or unavailable.
Finally, in terms of \textbf{Quality} dynamic methods generally produce higher-quality negative samples. Model-guided sampling delivers the highest quality by generating semantically meaningful negatives aligned with the evolving embedding space. Knowledge-constrained methods also generate high-quality samples by using ontologies and type constraints. Self-adaptive sampling though efficient, tends to produce lower-quality samples due to the absence of external knowledge, but it still adapts to the evolving embedding space better than static methods.

\subsection{Adversarial NS}
Adversarial negative sampling methods have gained significant attention, influenced by the performance of generative adversarial networks (GANs)~\cite{Goodfellow2014GenerativeAN}. In GANs, a generator creates counterfeit positive instances, while a discriminator differentiates between these and genuine positives. This process can be expressed as:
\begin{equation*}
    \underset{G}{min} \; \underset{D}{max}(\mathbb{E}_{P_{d}(y^+)}[log D(y^+)]+\mathbb{E}_{P_{G}(y^-)}[log(1-D(G(y^-)))]), 
\end{equation*}
where $G$ and $D$ are generator and discriminator respectively, $P_d$ is the positive data distribution and $P_G$ is the negative distribution generated by $G$.

\begin{figure}[]
\centering
    \includegraphics[width=0.5\textwidth]{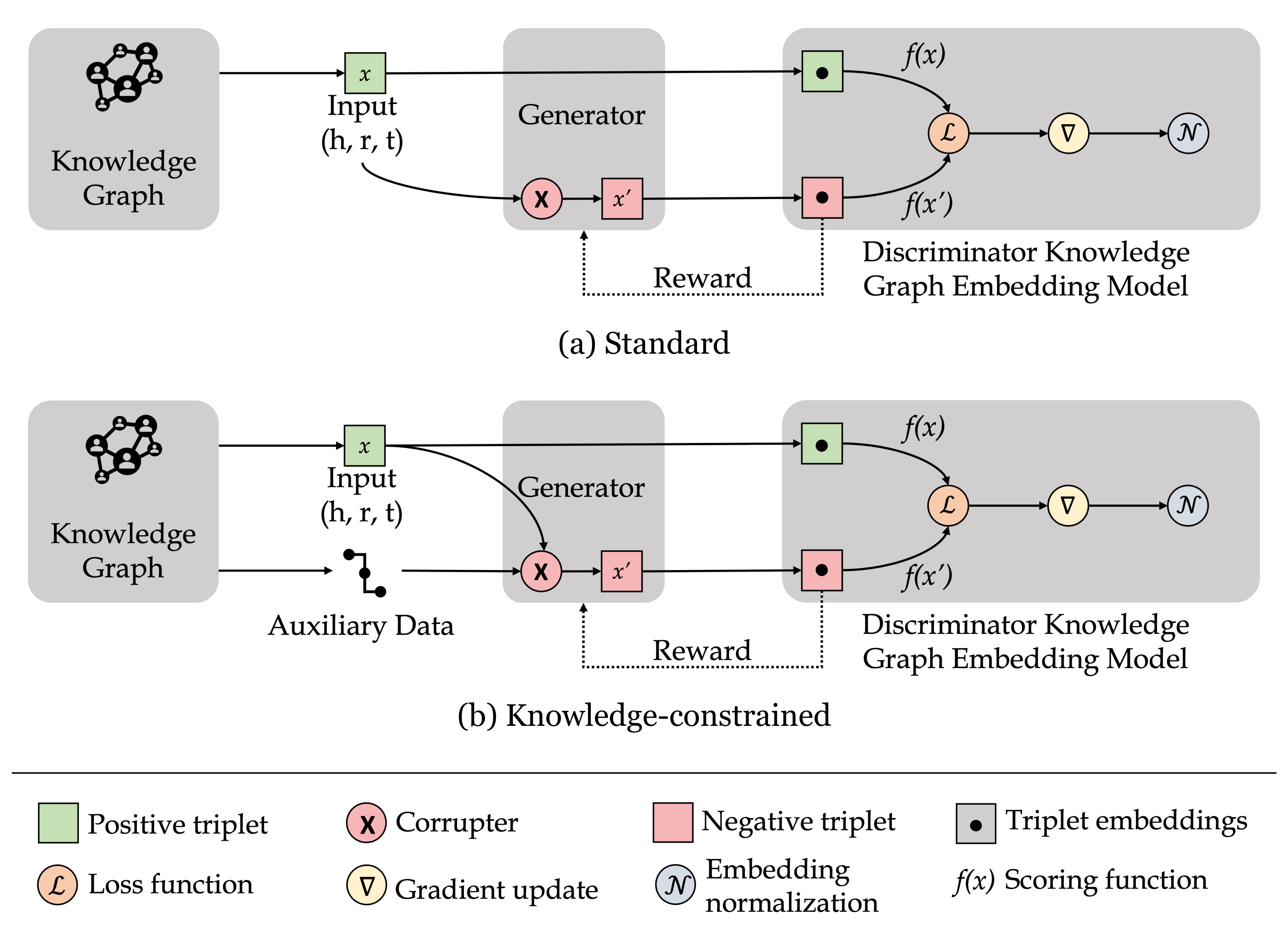}
\caption {Overview of adversarial negative sampling method and steps of training a knowledge graph embedding model over a positive instance $x=(h, r, t)$ and a corrupted negative $x'$ from an adversarial negative sampling method.}
\label{fig:adversarial_ns}
\end{figure}

In KGRL, the generator samples high-quality negative samples to deceive the discriminator. The goal is to generate hard negative samples, as illustrated in Figure~\ref{fig:adversarial_ns} and Adversarial methods can be categorized into two main types as below.
\begin{itemize}
    \item \textit{Standard}: This approach utilizes traditional GAN architectures to generate negative samples by training a generator to produce counterfeit negative instances that the discriminator cannot distinguish from positive ones. As shown in Figure~\ref{fig:adversarial_ns}(a), this iterative process enhances the quality of negative samples by continuously optimizing the generator based on the discriminator's feedback.
    \item \textit{Knowledge-constrained}: This variant enhances the GAN framework by integrating external knowledge sources to inform the generation of negative samples. As illustrated in Figure~\ref{fig:adversarial_ns}(b), these methods leverage additional contextual information, such as neighborhood relationships or semantic features, allowing the generator to produce more relevant and contextually accurate negative samples while improving the discriminator's evaluation capabilities.
\end{itemize}

\subsubsection{Standard}
In a pioneering application of Generative Adversarial Networks (GANs) to KGRL, KBGAN~\cite{Cai2018KBGANAL} utilizes policy gradient-based reinforcement learning for negative sampling, effectively confounding the discriminator. The generator, a KGE model, approximates the score distribution of candidate negative triples, presenting them in discrete form. The discriminator, serving as the target KGE model, evaluates the quality of the generated data.

IGAN (Incorporating GAN)~\cite{Wang2018IncorporatingGF} enhances adversarial negative sampling by employing a two-layer fully connected neural network as a generator, replacing the log-loss probability-based generator of KBGAN. The discriminator retains its role as the margin-based ranking loss KGE model. The generator network processes corrupted triplet embedding vectors through a non-linear activation function (\textit{ReLU}), followed by the \textit{softmax} function, to compute the probability distribution over the entire entity set.

GraphGAN~\cite{Wang2018GraphGAN} introduces a novel generator implementation termed \textit{Graph Softmax}, which addresses the shortcomings of traditional \textit{softmax} functions by considering graph structure and proximity. \textit{Graph Softmax} defines a connectivity distribution that ensures normalization, awareness of the underlying graph structure, and computational efficiency. An online generating strategy based on random walks is introduced, where the relevance probability of a neighbor entity $e_i$ given $e$ is formulated as $p_c(e_i \mid e) = \frac{exp (g_{e_i}^\top g_e)}{\sum_{e_j \in \mathcal{N}_c} exp (g_{e_j}^\top g_e)}$,
with $\mathcal{N}c(v)$ being the set of neighbors of entity $e$ and $g_e, g_{e_i} \in \mathbb{R}^k$ denoting the $k$-dimensional representation vectors. The \textit{graph softmax} calculates the probability of reaching each entity $e$ and $e$ from a root  $e_c$ as: $G(e \mid e_c) = (\prod_{j=1}^{m} p_c(e_{r_j} \mid e_{r_{j-1}}) . p_c(e_{r_{m-1}} \mid e_{r_m})$.
The generator approximates the connectivity distribution by adjusting the scores of its generated samples. Conversely, the discriminator is modeled as a \textit{sigmoid} function aimed at maximizing the log-probability of correctly classifying both positive and negative samples: $D(e,e_c) = \sigma(\textbf{d}_e^\top \textbf{d}_{e_c}) = \frac{1}{1 + exp(-\textbf{d}_e^\top \textbf{d}_{e_c})}$,
where $\textbf{d}_e, \textbf{d}_{e_c} \in \mathbb{R}^k$ are the $k$-dimensional representation vectors of entities $e$ and $e_c$.

KCGAN (Knowledge Completion GAN)~\cite{Zia2021KCGAN} advances the GAN-based approach for generative link prediction models, utilizing a game-theoretic framework that involves both a generator and a discriminator. This framework aims to enhance understanding of the underlying Knowledge Base (KB) structure by facilitating predicate/triplet classification and link prediction, thereby improving knowledge representation and completion capabilities. Unlike prior GAN-based methods in KGRL, where the generator constructs complete negative samples, KCGAN focuses on directly learning the link prediction task or the distribution over links. The discriminator reinforces this learned distribution by identifying positive samples.

The GN+DN (Generative Network + Discriminative Network)~\cite{LIU2022GNDN} method represents an extension of the GAN approach in KGRL, designed to generate previously unseen yet plausible instances. In this framework, the generator (GN) serves as the target KGE model for the knowledge graph completion task. It takes the vector representations of a head entity $h$ and a relation $r$ as input, transforming these to generate a feature representation of a potentially valid tail entity $\bar{t}_g$. The discriminator (DN) receives the generated triple $(h, r, \bar{t}_g)$, the ground truth triple $(h, r, t)$, and a randomly chosen negative sample $(h,r, \bar{t})$to compute the loss function, which distinguishes the true triple from other samples.

\subsubsection{Knowledge-constrained}
KSGAN (Knowledge Selective GAN)~\cite{Kairong2019KSGAN} extends KBGAN by introducing a knowledge selection step in the generator. This step filters out implausible triples and selects semantically relevant negative triples for a positive triplet. The primary goal remains to train the discriminator using negative triples produced by the generator, akin to KBGAN.

NKSGAN (Neighborhood Knowledge Selective GAN)~\cite{Hai2020NKSGAN} further enhances KSGAN by incorporating a neighbor aggregator component to generate high-quality negative samples and improve discriminator performance. This method aggregates neighborhood information to capture the semantic knowledge of entities, utilizing a graph attention mechanism to derive entity representations from a pre-trained KGE model with fixed parameters.

RUGA (Rules and Graph Adversarial learning)~\cite{Caifang2021RUGA} introduces a novel approach to enhance knowledge graph completion tasks. Initially, the conventional training of the KGE model is employed as both generator and discriminator, utilizing adversarial learning techniques to acquire high-quality negative samples. The generated negative samples, combined with existing positive samples, form labeled triples within the injection rule model. This method integrates triple information from the knowledge graph with additional logic rules, facilitating iterative learning through labeled triples, unlabeled triples, and soft rules. Each iteration involves alternating between the soft label prediction and embedding rectification stages.

NoiGAN~\cite{Cheng2020NoiGAN} is designed to learn noise-aware KGE by jointly training embedding and error detection tasks within a unified adversarial learning framework. It comprises two components: (i) a noise-aware KGE model that incorporates confidence scores $\mathcal{C}(h, r, t)$ to mitigate the noise impact on embeddings, generating high-quality embeddings and reliable correct triples $(h, r, t)$ for the adversarial model; (ii) a GAN for noise identification, which includes a noise generator that corrupts correct triples to produce probable noise, yielding reliable noise samples for the discriminator and high-quality negative samples for the KGE model. The discriminator assigns confidence scores to observed triples, effectively distinguishing correct triples from noise.

NAGAN-ConvKB~\cite{Le2021NAGAN-ConvKB} combines two NoiGAN models with a ConvKB model, utilizing the NoiGAN model to generate superior embedding vectors. This model consists of three primary components: the vector embedding model TransE, which creates embedding vectors representing the graph; the GAN model, responsible for selecting negative samples close to positive ones and calculating confidence scores; and finally, the ConvKB model, which uses the trained embedding vectors for link prediction. The GAN's input consists of the 
k\% embedding vectors that closely align with the model, i.e., achieving a minimum $||h + r - t||_1$ with $\mathcal{N}(h, r, t)$, the set of negative triples. Post-GAN training, the discriminator evaluates the confidence of true positive samples’ vectors, resulting in a binary value or a score within $[0,1]$. The generator assesses the proximity of negative samples to the positive sample, selecting $k$ negative samples for training the TransE model.

\subsubsection{Pros and Cons.}
When comparing standard GAN-based and Knowledge-constrained GAN methods for negative sampling in KGRL across five key aspects, several differences emerge. In terms of \textbf{Efficiency}, standard GAN methods are generally faster as they rely solely on internal reinforcement learning techniques, though their dual-model structure increases computational complexity. Knowledge-constrained methods introduce additional preprocessing steps that can increase time and space complexity due to external knowledge integration. 
For \textbf{Effectiveness}, standard GANs improve progressively by generating more challenging negative samples through the adversarial framework, while Knowledge-constrained methods benefit from the contextual relevance of external data, producing more accurate and semantically informed negative samples. However, the latter may suffer from bias if the external data is flawed. 
In terms of \textbf{Stability}, standard GANs are prone to instability, including mode collapse and oscillatory behavior, especially given the discrete nature of knowledge graphs. Knowledge-constrained methods generally offer more stability by using external context to guide training, though this stability hinges on the quality of the external data. 
\textbf{Adaptability} favors standard GANs, as they are more flexible and applicable across a range of knowledge graphs without requiring domain-specific information. Knowledge-constrained methods, while highly effective in domain-specific applications, are less adaptable when external information is unavailable or irrelevant, limiting their use in generalized contexts. 
Lastly, in terms of \textbf{Quality}, Knowledge-constrained methods excel at generating high-quality, contextually meaningful negative samples by leveraging external data, but they risk introducing biases if the external knowledge is incomplete or skewed. Standard GANs, although capable of generating difficult negative samples, sometimes produce less informative or trivial examples, particularly when the generator underperforms.

\subsection{Self Adversarial NS}

Self-adversarial negative sampling techniques have recently emerged as a novel and highly effective approach for generating negative triples in KGE models. These techniques leverage the current distribution of the target embedding model to sample negative triples based on the conditional probability distribution: 

\begin{equation*}
    P(\bar{x} \mid \{\bar{x}_n\}) = \frac{exp \; \alpha f(\bar{x})}{\sum exp \; \alpha f(\bar{x}_n)}, 
\end{equation*}
where $f(x)$ is the score function of the model, and $\bar{x}$ represents a candidate negative triple. The following are the three key subcategories of self-adversarial negative sampling approaches as illustrated in Figure \ref{fig:self_adv_ns}:

\begin{figure}[]
\centering
    \includegraphics[width=0.5 \textwidth]{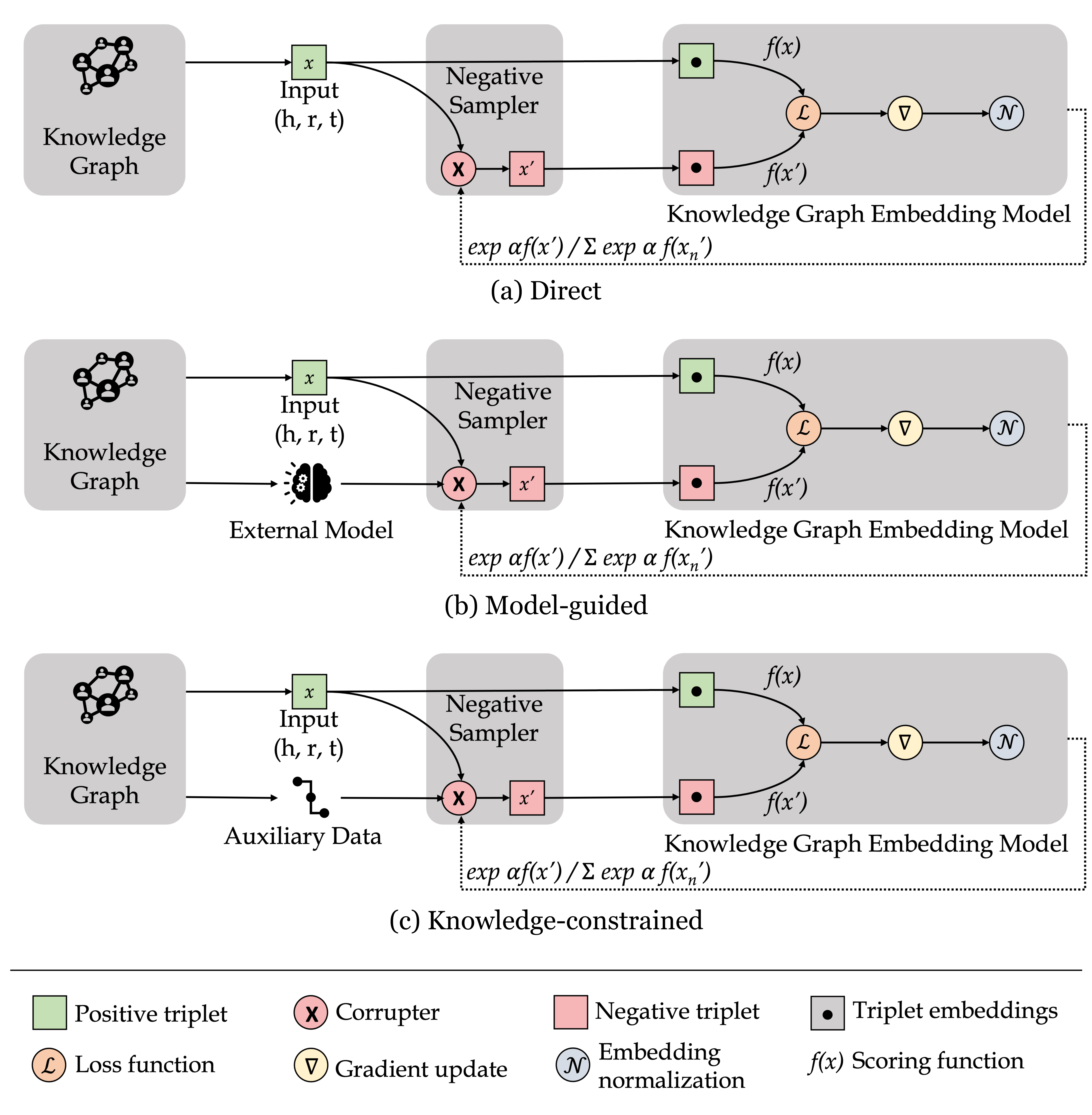}
\caption {Overview of subcategories of self-adversarial negative sampling methods and the steps of training a knowledge graph embedding model over a positive instance $x=(h, r, t)$ and a corrupted negative $x'$ from different self-adversarial negative sampling methods.}
\label{fig:self_adv_ns}
\end{figure}

\begin{itemize}
    \item \textit{Direct}: 
    This approach allows the model to dynamically generate negative samples based on its own learned distribution of negatives, adapting the selection process as the model evolves during training as depicted in Figure \ref{fig:self_adv_ns}(a). These methods directly exploit the model's internal structure to identify hard negative samples, encouraging the model to focus on the most challenging and informative negatives.
    \item \textit{Model-guided}:
    Model-guided methods enhance the self-adversarial framework by incorporating external or pre-trained models as illustrated in Figure \ref{fig:self_adv_ns}(b). These models generate more difficult or semantically meaningful negative samples, pushing the main learning model further by introducing greater variability and complexity into the training process.
    \item \textit{Knowledge-constrained}:
     This subcategory integrates external data sources, such as domain knowledge, schema information, and type constraints, into the adversarial negative sampling process as shown in Figure \ref{fig:self_adv_ns}(c). By incorporating auxiliary information, these methods can generate more refined and contextually meaningful negative samples, thus enhancing the model’s ability to learn richer representations.
\end{itemize}

\subsubsection{Direct}
CANS (Confidence-Aware Negative Sampling)~\cite{Shan2018ConfidenceAware} is a pivotal contribution to KGRL, introducing the concept of using the scoring function of the target KGE model to assess the confidence of generated negative samples. The quality of negative triples is quantified as $NQ(\bar{h}, r, \bar{t}) = -f(\bar{h}, r, \bar{t})$, with higher values indicating better quality. The confidence of each negative triple is computed using the \textit{softmax} function, allowing a probability distribution over a set of candidate negative triples. Specifically, the confidence of a negative triple $(\bar{h}, r, \bar{t})$ is defined as $NC(\bar{h}, r, \bar{t}) = \frac{exp \; NQ(\bar{h}, r, \bar{t})}{\sum_{(\bar{h}, r, \bar{t}) \in T_{NG}} exp \; NQ(\bar{h_i}, r_i, \bar{t_i})}$, where $T_{NG}$ is the set of candidate negative triples, and $\bar{h}$ and $\bar{t}$ are randomly selected from the entire entity set $\mathcal{E}$.

Self-Adv (Self-Adversarial)~\cite{Sun2019RotatEKG} applies the target KGE model to assess the difficulty of negative samples by sampling negatives from the distribution $p(\bar{h}_j,r,\bar{t}_j\mid \{(h_i,r_i,t_i)\}) = \frac{exp \; \alpha f(\bar{h}_j, r, \bar{t}_j)}{\sum_i exp \; \alpha f(\bar{h}_i, r, \bar{t}_i)}$, where $\alpha$ controls the temperature of sampling. Samples with higher gradients are considered hard negatives. However, this method risks generating false negatives, leading to performance overhead during gradient evaluation.

NSCaching~\cite{Zhang2019NSCachingSA} employs a cache mechanism to store negative samples for head and tail entities, updated asynchronously. This allows a larger set of candidates to be involved during training. Despite its benefits, updating probabilities of all samples remains computationally expensive.

LAS (Loss Adaptive Sampling)~\cite{Lei2019LAS} addresses false negatives in Self-Adv by using an adaptive probability mechanism based on loss functions. It incorporates a ``push-up" mechanism that prioritizes likely false negatives over true negatives. To limit the influence of false negatives from unrelated domains, LAS defines domains as observed head or tail domains of the relation and applies domain constraints to limit false negatives. Specifically, for a corrupted triple $(\bar{h}, r, t)$, LAS pushes it up only if $\bar{h} \notin \cup ; \textit{disDoms}(dom_r^{Head}, Doms)$, where $dom_r^{Head}$ is the observed head domain, $Doms$ is the set of domains of all relations, and $\textit{disDoms}(dom_r^{Head}, Doms)$ represents disjoint domains.

ASA (Adaptive Self-Adversarial)~\cite{qin2021ASA} addresses false negatives by focusing on moderately difficult samples rather than the hardest ones. It introduces a margin $\mu$ to control difficulty, where higher values correspond to easier samples. This reduces the chance of generating false negatives by ensuring that negative samples do not exceed the score of the positive relationships they originate from.

ESNS (Entity Similarity-based Negative Sampling)~\cite{Yao2022ESNS} introduces a shift-based logistic loss function for generating negative samples based on entity similarity, capturing both structural and semantic contexts. Entity similarity is quantified as Head Similarity ($S_h(e_i, e_j) = |C_h(e_i) \cap C_h(e_j)|$) and Tail Similarity ($S_t(e_i, e_j) = |C_t(e_i) \cap C_t(e_j)|$), where $C_h(e)$ ($C_h(e) = \{(r, t)\mid(e, r, t) \in F\}$) and $C_t(e)$ ($C_t(e) = \{(r, h)\mid(h, r, e) \in F\}$) are the structural contexts. The method also incorporates an entity inverted index $EII_{h/t}$ to store entity similarities and modifies the loss function in RotatE~\cite{Sun2019RotatEKG} to ensure higher-quality negative samples.

TANS (Triplet Adaptive Negative Sampling)~\cite{Feng2024TANS} extends Self-Adv methods by incorporating joint probability in the loss function, smoothing frequencies through model-predicted distributions. This approach provides a unified view of Self-Adv loss and subsampling, addressing data sparsity in negative sampling loss for KGE.

\subsubsection{Model-guided}
Integrating external models with Self-Adv has shown notable improvements in negative sampling for KGE models. Self-Adv SANS~\cite{Ahrabian2020StructureAN} builds on Self-Adv by using random walks to identify negatives, but it may generate low-quality samples when ignoring non-semantically similar neighbors. Self-Adv EANS~\cite{Je2022EntityAN} further combines entity-aware negative sampling with Self-Adv methods.

MCNS (Markov Chain Negative Sampling)~\cite{Yang2020underneg} approximates positive distributions using self-contrast approximation and accelerates sampling with Metropolis-Hastings. It employs depth-first search to traverse the graph and generate negatives, minimizing hinge loss for updating embeddings.

MDNCaching~\cite{tiroshan2022mdncaching} and TuckerDNCaching~\cite{tiroshan2022tuckerdncaching} use matrix decomposition-based latent relation models and caching mechanisms to preserve high-quality negative samples while addressing false negatives. The Tucker decomposition enhances semantic coherence in latent relation modeling.

ABNS (Attention-Based Negative Sampling)~\cite{si2023attention} introduces attention mechanisms to prioritize hard negatives while reducing the influence of false negatives. Two variants, ABNS-N (using zero-mean normalization) and ABNS-R (using \textit{ReLU} activation), compute attention scores for negative triplets. ABNS-N normalizes the deviation from the mean, while ABNS-R uses a \textit{ReLU} function to emphasize hard negatives. Over time, the method reduces attention on high-score negatives, maintaining the representation capability of KGE models by balancing attention between false and hard negatives.

CCS (Cluster-Cache Sampling)~\cite{Han2023CCS} generates negatives using entity similarity clustering. After partitioning the entity set into clusters using K-Means, it calculates triplet probabilities within caches and selects high-ranking candidates.

\subsubsection{Auxiliary data-based }
Local-cognitive NS~\cite{huang2020rate} combines Type-Constraints NS~\cite{Krompas2015TypeNS} with Self-Adv to introduce prior knowledge. Type-Constraints NS selects negative candidates by imposing relation-specific constraints but may suffer from graph sparsity. A dynamic coefficient $\gamma$ balances negative samples from constrained and unconstrained entity sets.

CAKE (Commonsense-Aware Knowledge Embedding)~\cite{Niu2022cake} enhances negative sampling by extracting commonsense knowledge from factual triples. Its three-module framework, i.e., Automatic Commonsense Generation, Commonsense-Aware negative sampling, and Multi-View Link Prediction, improves self-supervision and reduces false negatives, particularly for corrupted tail entities.

HaSa (Hardness and Structure-aware)~\cite{Zhang2024HaSa} addresses false negatives in KGE while preserving hard negative samples by introducing structural information. It modifies the InfoNCE loss and uses graph structure to differentiate between true and false negatives, leveraging shortest path lengths between entities.

\subsubsection{Pros and Cons.}
When comparing the subcategories of self-adversarial methods across the five aspects of negative sampling, distinct advantages and disadvantages emerge. \textbf{Efficiency} varies significantly among the methods; Direct methods achieve quick sampling using the model's scoring function, although they can incur high computational costs when updating probabilities for all negative candidates. Conversely, Model-guided methods increase the complexity of handling external models and large KG graphs. Knowledge-constrained methods excel in efficiency by limiting the search space with external knowledge, though they may become less efficient when accessing that knowledge. Notably, lazy update mechanisms in approaches like NSCaching, and CCS significantly enhance efficiency, especially in large-scale settings.
In terms of \textbf{Effectiveness}, Model-guided methods tend to excel by utilizing external structures to create more relevant hard negatives, thereby improving training outcomes. Direct methods are effective at generating challenging samples but run the risk of introducing false negatives, which can hinder model performance. Knowledge-constrained methods also offer high effectiveness, leveraging prior knowledge to filter out irrelevant negatives, yet their reliance on external resources may limit flexibility. 
\textbf{Stability} is generally a strong suit for Knowledge-constrained methods, as their structured approach results in consistent performance across datasets. Direct methods demonstrate superior stability compared to model-guided methods, but they still risk instability due to the potential for generating false negatives. Meanwhile, Model-guided methods may exhibit variable stability based on the reliability of the external models they incorporate, making their performance more contingent on additional resources. 
Regarding \textbf{Adaptability}, Model-guided methods excel due to their flexibility in working with diverse KGE models and graph structures, although they may struggle in sparsely connected graphs. Direct methods are also adaptable but may not handle complex knowledge graph semantics as well as their model-guided counterparts. Knowledge-constrained methods are adaptable in domains with available external knowledge but may become limited when such resources are inaccessible.
Finally, concerning \textbf{Quality}, Knowledge-constrained methods produce high-quality negative samples by ensuring semantic relevance through external knowledge. Model-guided methods also prioritize quality by emphasizing hard negatives, but they risk decreasing quality over time if not managed properly. Direct methods can achieve quality by generating challenging negatives but may include false negatives that undermine overall training efficacy. The issue of false negatives represents a significant challenge, as the incomplete nature of knowledge graphs often leads to the inclusion of false negative triples during the training process, potentially misleading the model. 

\subsection{Mix-up NS}

Mixup~\cite{Hongyi2018Mixup} is a data augmentation technique that constructs new data points by creating convex combinations of sample pairs, thereby promoting linear behavior among training samples. This method is predominantly employed in supervised learning and negative sampling. Mixup generates artificial training samples and their corresponding labels in supervised learning through linear interpolation. In contrast, within negative sampling, the mixing operation facilitates the creation of challenging negative samples by linearly interpolating virtual samples. Architectures of mixing negative sampling techniques are illustrated in Figure~\ref{fig:mixing_ns}.

\begin{figure}[]
\centering
    \includegraphics[width=0.5\textwidth]{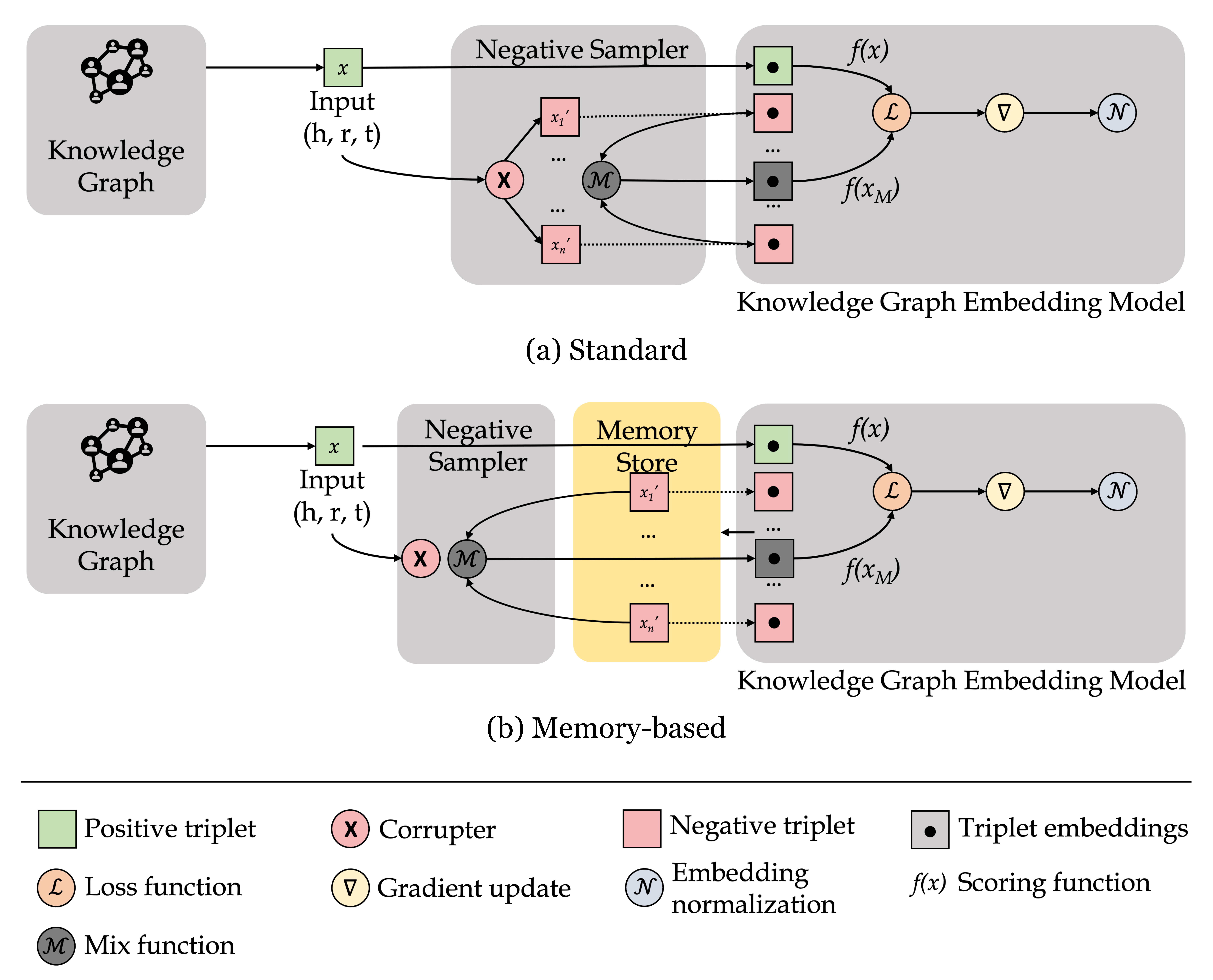}
\caption {Overview of mixing negative sampling methods and steps of training a knowledge graph embedding model over a positive instance $x=(h, r, t)$ and a new negative $x'$ generated by mixing candidate negatives $\{x'_1, ..., x'_n\}$ from different mix-up negative sampling methods.}
\label{fig:mixing_ns}
\end{figure}

The Mix-Up approach in KGRL can be considered with two main subcategories as follows:
\begin{itemize}
    \item \textit{Standard}: This subcategory focuses on generating negative samples through the linear interpolation of entities within the knowledge graph. It emphasizes the identification of hard negative triplets, combined to create more challenging samples as shown in Figure~\ref{fig:mixing_ns}(a). 
    \item \textit{Memory-based}: This subcategory employs a memory-efficient framework for generating high-quality negative samples. It utilizes a cache system that dynamically stores samples from high-scoring negative triples as illustrated in Figure~\ref{fig:mixing_ns}(b). By incorporating a mixing operation that combines entities from different caches, these approaches create more challenging negative samples while balancing the need for reliable hard negatives.
\end{itemize}

\subsubsection{Standard}
MixKG~\cite{Che2022MixKGMF} introduces a framework for negative sampling in KGRL, employing the mixup technique to construct negative triplets involving non-existent entities in knowledge graphs. It first identifies hard negative triplets through two selection criteria: a score function-based selector (HNS-SF) and a correct entity similarity selector (HNS-CES). MixKG selects $k$ negative triplets from each HNS-SF, prioritizing candidates with high scores as difficult negatives, while HNS-CES focuses on negative entities similar to the positive entity. These hard negatives are mixed to generate even more challenging samples. For example, given two negative triplets $(h, r, \bar{t}_i)$ and $(h, r, \bar{t}_j)$, the tail entity of the new, more difficult negative triplet is computed as: $\hat{\textbf{t}}_{i,j} = \alpha \cdot \bar{\textbf{t}}_i + (1-\alpha) \cdot \bar{\textbf{t}}_j$, where $\alpha$ is a randomly selected value in $(0, 1)$. The resulting triplet is $(h, r, \hat{t}_{i,j})$.

Many existing methods assume that high-scoring, non-existent triples are reliable negative samples. However, this may introduce noise since some triples, known as false negatives, could represent true facts due to the incompleteness of knowledge graphs. DeMix~\cite{Chen2023DeMix} addresses this issue by proposing a denoising mixup approach that improves negative sample quality through self-supervised refinement. DeMix operates with two key modules: the Marginal Pseudo-Negative Triple Estimator (MPNE) and Adaptive Mixup (AdaMix). MPNE classifies corrupted triples (e.g., $(h,r,e)$ or $(e,r,t)$) into marginal pseudo-negatives or true negatives using current model outputs. AdaMix selects appropriate mixup partners for the entities, generating partially positive triples for pseudo-negatives and harder triples for true negatives, thereby enhancing training data quality and improving KGE performance.

M²ixKG~\cite{Che2024M2ixKG} further enhances KGE by employing mixup in two ways: (1) mixing head and tail entities within triplets sharing the same relation to generate synthetic triplets, and (2) mixing high-scoring negative triplets to create more difficult samples. For triplets with identical relations, new synthetic triplets are formed by interpolating head and tail entities as $h_{ij} = \alpha_1 \cdot h_i + (1 - \alpha_1) \cdot h_j$, and $\quad t_{ij} = \alpha_1 \cdot t_i + (1 - \alpha_1) \cdot t_j$, where $\alpha_1$ is sampled from a beta distribution $\alpha_1 \in (0, 1)$. These synthetic triplets $(h_{ij}, r, t_{ij})$ are filtered based on scores, and only those exceeding a defined margin are used in training. The second phase of M²ixKG introduces a flexible selection mechanism for hard negatives, dynamically sampling them based on a probability distribution proportional to their scores. For each selected hard negative pair $(h, r, t'_i)$ and $(h, r, t'_j)$, M²ixKG mixes the tail entities, yielding a harder triplet: $\hat{t}_{ij} = \alpha_2 \cdot t'_i + (1 - \alpha_2) \cdot t'_j$, where $\alpha_2$ is also sampled from a beta distribution. The resulting triplet $(h, r, \hat{t}_{ij})$ serves as a more challenging negative candidate, improving model training.

\subsubsection{Memory-based }
The DCNS~\cite{Zheng2024DCNS} (Double-Cache Negative Sampling) method offers an innovative strategy for generating high-quality negative samples in knowledge graph embedding. It utilizes a dual-cache system, $\mathcal{H}(r,t)$ and $\mathcal{T}(h,r)$, to store and sample from high-quality negatives, addressing the vanishing gradient issue. DCNS dynamically updates these caches during training to keep them relevant as the data distribution evolves. It generates harder negatives by mixing samples from different caches using an adaptive weight, $\bar{h}_{1,2} = \alpha * \bar{h}_1 +(1-\alpha) * \bar{h}_2$ where $\bar{h}_1$ is drawn from $\mathcal{H}^1(r,t)$ and $\bar{h}_2$ from $\mathcal{H}^2(r,t)$. The sampling strategy selects the top $K$ highest-scoring samples from each cache, balancing the need for hard negatives with the risk of false negatives. By drawing multiple samples per round, DCNS further reduces the potential impact of false negatives.
\subsubsection{Pros and Cons.}
In terms of \textbf{Efficiency}, standard mix-up methods typically involve straightforward linear operations that are computationally light, but can become costly in large-scale graphs when generating numerous mixed negatives. On the other hand, memory-based methods improve efficiency by utilizing a cache system that reduces redundant sampling efforts, though at the cost of increased memory consumption to maintain and update the caches.
For \textbf{Effectiveness}, standard methods focus on generating challenging negatives by combining hard negatives through interpolation, which drives model learning. However, their effectiveness can depend on the quality of the initial negative samples. Memory-based methods maintain high effectiveness by dynamically updating the cache with high-quality negatives, ensuring a continuously improving negative pool that increases the difficulty of samples over time.
Regarding \textbf{Stability}, standard mix-up approaches generally perform well across different datasets, but they may be sensitive to hyperparameters, leading to variability in results. Memory-based methods offer greater stability through the dynamic update of caches, which ensures relevant samples are retained.
In terms of \textbf{Adaptability}, standard mix-up methods are flexible and can be applied to a variety of knowledge graphs and embedding models without needing extensive assumptions. However, memory-based methods show higher adaptability as the caching mechanism is more generic, allowing it to store diverse negative samples from various graph structures.
Finally, for \textbf{Quality}, standard mix-up methods often generate high-quality negatives, especially with models like DeMix that address false negatives, but may suffer if the original negatives are not sufficiently hard. Memory-based methods excel in maintaining high-quality negatives, as the cache system ensures only challenging and informative negatives are retained and mixed, although they might still introduce noise if false negatives are stored in the cache. Commonly in both mix-up methods, the presence of non-existent entities within knowledge graphs poses a challenge, as it may result in the generation of semantically meaningless negatives that could potentially deceive the KGE model.

\subsection{Text-based NS}
Text-based negative sampling leverages textual information from entities and relations to generate semantically meaningful negative samples, enhancing representation learning. Negatives are sampled from the same decoding distribution as the anchor, making them semantically close and thus ``hard" negatives. Sequence-to-sequence models generate tail entities given a head entity and relation pair, with incorrect entities serving as negatives. This approach improves semantic similarity, as negatives are derived from the same hidden state as the correct fact. Techniques such as self-information-enhanced contrastive learning can also be employed to increase the diversity of negatives. During inference, these hard negatives act as high-quality candidates, enabling faster selection of entities in large-scale knowledge graphs. The architecture of text-based methods is illustrated in Figure~\ref{fig:text_ns}.

\begin{figure}[]
\centering
    \includegraphics[width=0.5\textwidth]{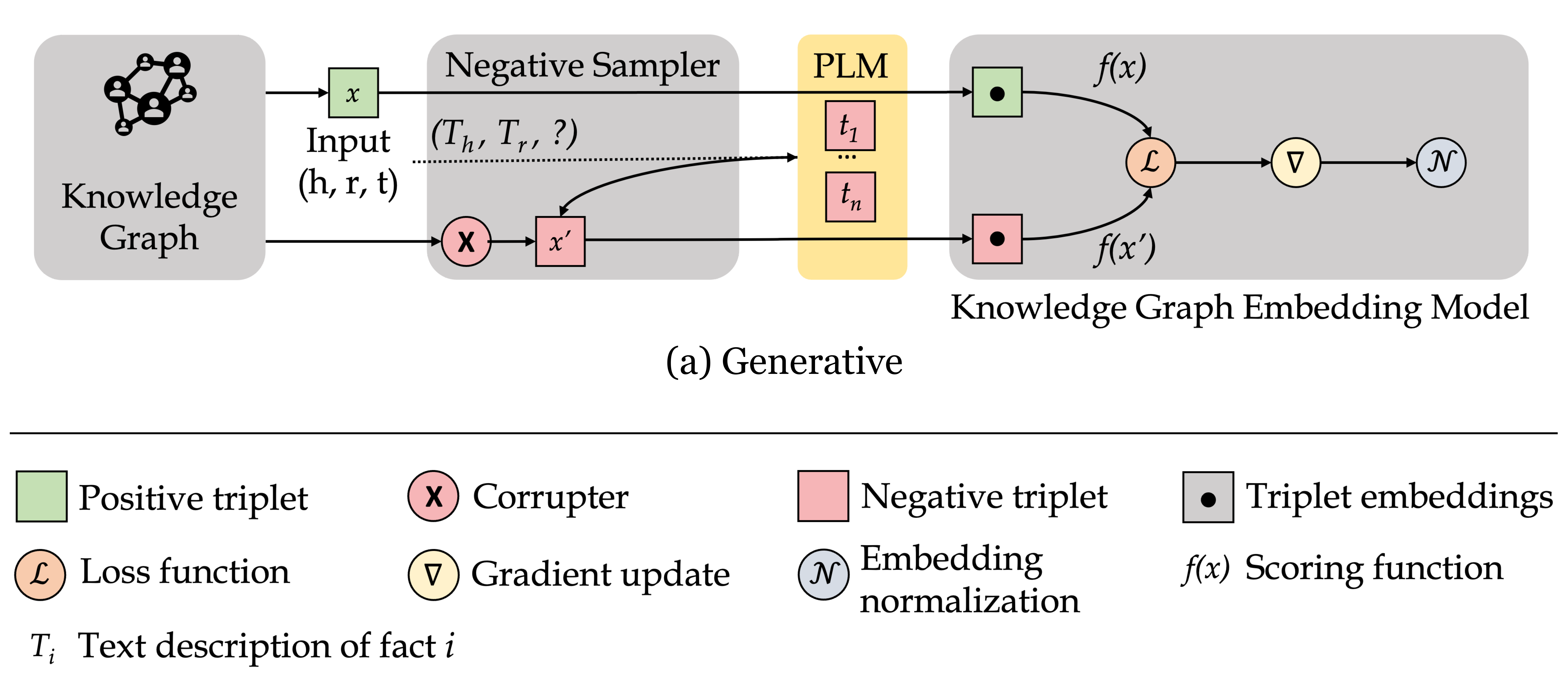}
\caption {Overview of mixing negative sampling method and steps of training a knowledge graph embedding model over a positive instance $x=(h, r, t)$ and a corrupted candidate negatives $x'$ from generative text-based negative sampling method.}
\label{fig:text_ns}
\end{figure}

The Text-based NS approach in KGRL introduces a novel contribution under the following sub-category.
\begin{itemize}
\item \textit{Generative}: Generative methods leverage generative models to create hard negative samples from head-relation pairs. Generating negatives that are semantically close to the correct entities enhances training effectiveness. Additionally, it improves inference efficiency by narrowing the candidate search space, making it a novel and effective approach for large-scale knowledge graph tasks. 
\end{itemize}

\subsubsection{Generative}

GHN (Generative Hard Negative)~\cite{Qiao2023GHN} employs a sequence-to-sequence model like BART~\cite{Lewis2020Bart} to generate tail entities from a head entity and relation pair. It uses a dual-encoder architecture: one encoder ($E_{hr}$) for the head entity and relation, and another ($E_t$) for the tail entity and its description. The resulting \textit{mean-pooled} and \textit{l2-normalized} embeddings are compared using \textit{cosine similarity}: $s(h, r, t) = cos(E_{hr}, E_t)$. GHN uses an InfoNCE loss function: $\mathcal{L} = -log(e^{s(h, r, t)/\tau} / e^{s(h, r, t)/\tau} + \Sigma e^{s(h, r, t_i)}/\tau)$, where $\tau$ is a learnable parameter. It further incorporates self-information-enhanced training for greater negative sample diversity. GHN's two-stage inference first generates a small set of candidate entities using the sequence-to-sequence model, followed by re-ranking using $s(h, r, t_i)$, improving inference speed in large-scale datasets while maintaining flexibility for smaller graphs.


%
\subsubsection{Pros and Cons.}
In terms of \textbf{Efficiency}, generative methods can introduce significant computational overhead due to the complexity of sequence-to-sequence models. However, they enhance inference speed through a two-stage process that narrows down the candidate search space. 
Regarding \textbf{Effectiveness}, these generative approaches excel at producing semantically meaningful hard negatives, improving representation learning by effectively challenging the model to differentiate between true and false triples.
\textbf{Stability} is generally high for generative methods, as they consistently generate meaningful negatives across diverse datasets, although their performance can vary based on the quality of textual descriptions. 
In terms of \textbf{Adaptability}, generative methods offer flexibility to work with different types of knowledge graphs but may struggle in scenarios where textual information is limited or of low quality.
Finally, in terms of \textbf{Quality}, generative text-based methods typically produce high-quality negatives that are semantically close to true facts, which enhances the model's ability to discern correct triples from incorrect ones. However, there remains a risk of generating false negatives, particularly if the textual information is ambiguous. 

%
\section{Conclusion and Open Research Directions}
\label{section:conclusion}
This paper provides a comprehensive review of negative sampling methods employed in knowledge graph representation learning. A summary of these methods can be found in Table \ref{table:ns}. To facilitate analysis and comparison, we categorize existing negative sampling techniques into six distinct categories, discussing the strengths and limitations of each approach. Considering the identified challenges and the advancements made in the field of negative sampling, we outline several promising research directions that can further advance this area of study.
\subsection{Mitigating False Negatives}
False negatives, i.e., instances where a true positive is incorrectly classified as a negative sample, represent a significant challenge in the development of effective negative sampling strategies for KGRL. The goal of these techniques is to generate semantically meaningful negative samples, yet the risk of false negatives can increase if proper mitigation strategies are not employed. While it is unrealistic to entirely eliminate false negatives, several innovative methods have been introduced to tackle this issue.

One such method is Entity Aware Negative Sampling~\cite{Je2022EntityAN}, which employs a novel scoring function to assess the likelihood of a triple, facilitating the differentiation between true and false negatives. Loss Adaptive Sampling~\cite{Lei2019LAS} incorporates a push-up mechanism to adjust sampling probabilities, thereby reducing the impact of false negatives. Similarly, the Adaptive Self-Adversarial approach~\cite{qin2021ASA} focuses on moderately difficult negative samples rather than exclusively on the most challenging ones. Other techniques, such as MDNCaching~\cite{tiroshan2022mdncaching} and TuckerDNCaching~\cite{tiroshan2022tuckerdncaching}, leverage pre-trained latent relation models to predict and filter potential false negatives. Reason-KGE~\cite{Jain2021ReasonKGE} utilizes an ontology-driven method that employs symbolic reasoning to ensure prediction consistency and improve the generation of reliable negative samples. Attention-Based Negative Sampling~\cite{si2023attention} applies attention mechanisms to balance the selection of hard and false negatives, refining training sample selection. Recently, DeMix~\cite{Chen2023DeMix} introduced a self-supervised refinement mechanism that classifies corrupted triples into marginal pseudo-negatives and true negatives to enhance the quality of training data.

Future research in the mitigation of false negatives in KGRL presents a broad array of opportunities aimed at improving the reliability of negative sampling techniques. One promising direction is the development of advanced modeling techniques that capture the complex relationships between entities, potentially leveraging context-aware models that incorporate surrounding entities and their semantic interactions. Incorporating external knowledge sources, such as ontologies, could further validate negative samples and enhance the accuracy of distinguishing between true and false negatives. Additionally, establishing quality metrics to assess the likelihood of false negatives would facilitate targeted filtration strategies, incorporating factors like semantic similarity, co-occurrence frequency, and contextual relevance. Balancing the reduction of false negatives with computational efficiency remains a critical area of focus, as more sophisticated techniques may introduce processing overhead. Developing efficient algorithms that maintain high-quality negative sampling while minimizing computational costs is essential for real-world applications. Enhancing model robustness against noisy or incomplete data through methods like noise filtering, outlier detection, and uncertainty quantification could significantly strengthen the model's ability to generate reliable negatives. Finally, employing ensemble approaches that combine multiple negative sampling methods may leverage the strengths of diverse techniques, leading to more resilient KGRL systems. Addressing these complexities surrounding false negatives is crucial for advancing the field of KGRL.
\subsection{Mix-up Negatives}
Mix-up approaches in KGRL have emerged as a promising strategy for enhancing the quality of negative sampling, which is essential for the development of robust and accurate knowledge graph embeddings. By integrating mix-up techniques, particularly those that utilize linear interpolation between hard negatives, researchers can generate more challenging and informative samples. This innovative method effectively addresses limitations commonly associated with traditional negative sampling approaches, such as the reliance on simplistic negatives and the risk of false negatives arising from the inherent incompleteness of knowledge graphs.

However, challenges remain in the application of mix-up techniques within KGRL. A significant concern is the computational overhead involved in dynamically generating and mixing negative samples, especially in large-scale knowledge graphs that encompass millions of entities and relations. Although methods like MixKG~\cite{Che2022MixKGMF} and M²ixKG~\cite{Che2024M2ixKG} demonstrate efficiency in producing synthetic samples, they often entail a trade-off between sample quality and computational cost. Furthermore, the success of mix-up methods is heavily dependent on the initial quality of negative samples; if these are poorly selected, the mixing process may fail to adequately mitigate noise during training. Additionally, the need for hyperparameter tuning, i.e., particularly regarding mixing coefficients and negative triplet selection mechanisms, poses another challenge, as the sensitivity of these approaches to configuration complicates their application in increasingly complex knowledge graphs.

Considering these challenges, future research could explore several promising avenues to advance mix-up techniques in KGRL. One direction involves the development of adaptive mix-up strategies that dynamically adjust mixing coefficients based on the model's current performance, allowing for a more responsive approach to generating hard negatives while minimizing false negatives. Another avenue is the design of scalable, memory-efficient methods that combine mix-up with memory-based techniques, such as DCNS~\cite{Zheng2024DCNS}, ensuring a balance between computational efficiency and the generation of high-quality negative samples. Addressing these open research questions could significantly enhance the applicability and performance of mix-up methods in the context of KGRL, ultimately leading to more effective knowledge graph embeddings.

\subsection{Text-based Negative Sampling}
Text-based negative sampling methods in KGRL have proven highly effective by utilizing textual information related to entities and relations, allowing for the generation of challenging and semantically meaningful negative samples. These samples help train models to distinguish between true and false triples more effectively, thereby improving the discriminative power of knowledge graph embeddings. For example, Generative Hard Negative sampling~\cite{Qiao2023GHN} uses sequence-to-sequence models such as BART~\cite{Lewis2020Bart} to generate hard negative samples by predicting tail entities given a head-relation pair. By employing a dual-encoder architecture that generates embeddings for both the input pair and its negative counterpart, GHN ensures the negative samples are semantically close to positive samples, which helps in refining model accuracy.

While models like GHN have advanced the field, challenges related to scalability and computational efficiency persist, particularly due to the complexity of generative architectures. These methods require substantial computational resources, limiting their applicability in real-time settings or large-scale systems. To address this, future research could focus on optimizing model architectures through techniques like pruning and distillation, which reduce computational overhead without sacrificing the quality of negative samples. Additionally, hybrid approaches that integrate text-based negative sampling with other methods could balance the trade-offs between sample diversity and computational efficiency. Open questions remain regarding how best to generalize these methods to scenarios where textual data is limited or noisy, as well as mitigating potential biases introduced by the use of such text-based strategies. The continued exploration of these areas offers a promising path toward more scalable and versatile KGRL systems.
\subsection{Nonnegative Sampling in KGRL}
To address the challenges associated with negative sampling, including potential instability in model performance, a promising avenue is to explore efficient non-sampling strategies or entirely eliminate negative samples in KGRL. Despite limited research in this area, several models have emerged that operate without negative samples, targeting the issues stemming from sample quality.

BKENE (Bootstrapped Knowledge Graph Embedding based on Neighbor Expansion)~\cite{Seon2022BootstrappedKGE} introduces a framework that generates knowledge graph representations without negative samples. It utilizes two encoders to create semantically similar views: one aggregates information from directly connected nodes and relations, while the other incorporates data from expanded multi-hop neighbors. This approach captures diverse relational information while minimizing semantic distortions typically caused by augmentation techniques.
The ``Stay Positive" (SP) method~\cite{hajimoradlou2022stay} eliminates negative sampling by incorporating a regularization term in the loss function, which encourages high scores for true triples while keeping the sum of scores across all triples constant. 

NS-KGE~\cite{Zelong2021NSKGE} circumvents negative sampling by dynamically weighting all negative instances in the knowledge graph. The model trains a scoring function to assess the appropriateness of linking head and tail entities through specific relations, minimizing discrepancies between ground truth and predicted values.
NSF-KGE~\cite{Bahaj2024NSFKGE} presents a framework that completely avoids negative sampling, effectively addressing the Closed World Assumption and its associated challenges. This model embeds positive facts and generates compositional embeddings from heads, tails, and relations, which are then optimized using a negative-sample-free loss function.
KG-NSF~\cite{bahaj2022kgnsf} also forgoes negative sampling by leveraging cross-correlation matrices of embedding vectors, thereby avoiding biases associated with traditional scoring functions. This method employs a BT (Barlow Twins) loss~\cite{Jure2021BTLoss} objective to maximize cross-correlation between positive pairs, enhancing the quality of learned embeddings.

Future research should focus on further developing non-sampling strategies to enhance KGRL, including scalability to large knowledge graphs, hybrid approaches that integrate sampling techniques, and exploring theoretical frameworks to better understand the implications of non-sampling methods on model performance and generalization.
%
\bibliographystyle{IEEEtran}
\bibliography{paper}

\begin{thebibliography}{100}
\providecommand{\url}[1]{#1}
\csname url@samestyle\endcsname
\providecommand{\newblock}{\relax}
\providecommand{\bibinfo}[2]{#2}
\providecommand{\BIBentrySTDinterwordspacing}{\spaceskip=0pt\relax}
\providecommand{\BIBentryALTinterwordstretchfactor}{4}
\providecommand{\BIBentryALTinterwordspacing}{\spaceskip=\fontdimen2\font plus
\BIBentryALTinterwordstretchfactor\fontdimen3\font minus \fontdimen4\font\relax}
\providecommand{\BIBforeignlanguage}[2]{{%
\expandafter\ifx\csname l@#1\endcsname\relax
\typeout{** WARNING: IEEEtran.bst: No hyphenation pattern has been}%
\typeout{** loaded for the language `#1'. Using the pattern for}%
\typeout{** the default language instead.}%
\else
\language=\csname l@#1\endcsname
\fi
#2}}
\providecommand{\BIBdecl}{\relax}
\BIBdecl

\bibitem{Carlson2010NELL}
A.~Carlson, J.~Betteridge, B.~Kisiel, B.~Settles, E.~R. Hruschka, and T.~M. Mitchell, ``Toward an architecture for never-ending language learning,'' in \emph{Proceedings of Twenty-Fourth AAAI Conference on Artificial Intelligence}, 2010, pp. 1306--1313.

\bibitem{Bollacker2008Freebase}
K.~D. Bollacker, C.~Evans, P.~K. Paritosh, T.~Sturge, and J.~Taylor, ``Freebase: a collaboratively created graph database for structuring human knowledge,'' in \emph{Proceedings of International Conference on Management of Data}, 2008, pp. 1247--1250.

\bibitem{Auer2007DBpediaAN}
S.~Auer, C.~Bizer, G.~Kobilarov, J.~Lehmann, R.~Cyganiak, and Z.~G. Ives, ``{DB}pedia: {A} nucleus for a web of open data,'' in \emph{Proceedings of International Semantic Web Conference}, vol. 4825, 2007, pp. 722--735.

\bibitem{Miller1995WordNet}
G.~A. Miller, ``{WordNet}: {A} lexical database for english,'' \emph{Communication of the ACM}, vol.~38, no.~11, pp. 39--41, 1995.

\bibitem{Suchanek2007Yago}
F.~M. Suchanek, G.~Kasneci, and G.~Weikum, ``Yago: a core of semantic knowledge,'' in \emph{Proceedings of Sixteenth International Conference on World Wide Web}, 2007, pp. 697--706.

\bibitem{Xiao2019QA}
X.~Huang, J.~Zhang, D.~Li, and P.~Li, ``Knowledge graph embedding based question answering,'' in \emph{Proceedings of Twelfth ACM International Conference on Web Search and Data Mining}, 2019, pp. 105--113.

\bibitem{Davide2016ExemplarQueries}
D.~Mottin, M.~Lissandrini, Y.~Velegrakis, and T.~Palpanas, ``Exemplar queries: A new way of searching,'' \emph{The International Journal on Very Large Data Bases}, vol.~25, no.~6, pp. 741--765, 2016.

\bibitem{Nararatwong2022FinQA}
R.~Nararatwong, N.~Kertkeidkachorn, and R.~Ichise, ``Enhancing financial table and text question answering with tabular graph and numerical reasoning,'' in \emph{Proceedings of Second Conference of the Asia-Pacific Chapter of the Association for Computational Linguistics and the Twelfth International Joint Conference on Natural Language Processing}, 2022, pp. 991--1000.

\bibitem{Xiang2019KGAT}
X.~Wang, X.~He, Y.~Cao, M.~Liu, and T.-S. Chua, ``{KGAT}: Knowledge graph attention network for recommendation,'' in \emph{Proceedings of Twenty-Fifth International Conference on Knowledge Discovery and Data Mining}, 2019, pp. 950--958.

\bibitem{Fuzheng2016KGRecommender}
F.~Zhang, N.~J. Yuan, D.~Lian, X.~Xie, and W.-Y. Ma, ``Collaborative knowledge base embedding for recommender systems,'' in \emph{Proceedings of Twenty-Second International Conference on Knowledge Discovery and Data Mining}, 2016, pp. 353--362.

\bibitem{Hongwei2018DKN}
H.~Wang, F.~Zhang, X.~Xie, and M.~Guo, ``{DKN}: Deep knowledge-aware network for news recommendation,'' in \emph{Proceedings of World Wide Web Conference}, 2018, pp. 1835--1844.

\bibitem{Venkatesh2022ConversationalIR}
P.~R. Venkatesh, K.~Chaitanya, R.~Kumar, and P.~R. Krishna, ``Conversational information retrieval using knowledge graphs,'' in \emph{Proceedings of Thirty-First ACM International Conference on Information and Knowledge Management and First Workshop on Proactive and Agent-Supported Information Retrieval}, 2022.

\bibitem{Dietz2018UtilizingKG}
L.~Dietz, A.~Kotov, and E.~Meij, ``Utilizing knowledge graphs for text-centric information retrieval,'' in \emph{Proceedings of Fourty-First International Conference on Research and Development in Information Retrieval}, 2018, pp. 1387--1390.

\bibitem{Dong2014Kvault}
X.~Dong, E.~Gabrilovich, G.~Heitz, W.~Horn, N.~Lao, K.~Murphy, T.~Strohmann, S.~Sun, and W.~Zhang, ``{K}nowledge {V}ault: A web-scale approach to probabilistic knowledge fusion,'' in \emph{Proceedings of Twentieth International Conference on Knowledge Discovery and Data Mining}, 2014, pp. 601--610.

\bibitem{Krompas2015TypeNS}
D.~Krompa{\ss}, S.~Baier, and V.~Tresp, ``Type-constrained representation learning in knowledge graphs,'' in \emph{Proceedings of International Semantic Web Conference}, 2015, pp. 640--655.

\bibitem{Moher2009PRISM}
D.~Moher, A.~Liberati, J.~Tetzlaff, and D.~G. Altman, ``Preferred reporting items for systematic reviews and meta-analyses: the {PRISMA} statement,'' \emph{British Medical Journal}, vol. 339, 2009.

\bibitem{Wang2021Survey}
M.~Wang, L.~Qiu, and X.~Wang, ``A survey on knowledge graph embeddings for link prediction,'' \emph{Symmetry}, vol.~13, no.~3, p. 485, 2021.

\bibitem{Ji2022Survey}
S.~Ji, S.~Pan, E.~Cambria, P.~Marttinen, and P.~S. Yu, ``A survey on knowledge graphs: Representation, acquisition, and applications,'' \emph{{IEEE} Transactions on Neural Networks and Learning Systems}, vol.~33, no.~2, pp. 494--514, 2022.

\bibitem{tiroshan2022tuckerdncaching}
T.~Madushanka and R.~Ichise, ``Tucker{DNC}aching: high-quality negative sampling with tucker decomposition,'' \emph{Journal of Intelligent Information Systems}, vol.~61, pp. 739--763, 2023.

\bibitem{Bordes2013Translating}
A.~Bordes, N.~Usunier, A.~Garcia-Dur\'{a}n, J.~Weston, and O.~Yakhnenko, ``Translating embeddings for modeling multi-relational data,'' in \emph{Proceedings of Thirteenth International Conference on Neural Information Processing Systems}, 2013, pp. 2787--2795.

\bibitem{Wang2014Knowledge}
Z.~Wang, J.~Zhang, J.~Feng, and Z.~Chen, ``Knowledge graph embedding by translating on hyperplanes,'' in \emph{Proceedings of Fourteenth AAAI Conference on Artificial Intelligence}, 2014, pp. 1112--1119.

\bibitem{Lin2015Learning}
Y.~Lin, Z.~Liu, M.~Sun, Y.~Liu, and X.~Zhu, ``Learning entity and relation embeddings for knowledge graph completion,'' in \emph{Proceedings of Twenty-Ninth AAAI Conference on Artificial Intelligence}, 2015, pp. 2181--2187.

\bibitem{Ji2015Knowledge}
G.~Ji, S.~He, L.~Xu, K.~Liu, and J.~Zhao, ``Knowledge graph embedding via dynamic mapping matrix,'' in \emph{Proceedings of Fifty-Third Annual Meeting of the Association for Computational Linguistics and Seventh International Joint Conference on Natural Language Processing)}, 2015, pp. 687--696.

\bibitem{Zhang2017TransHR}
C.~Zhang, M.~Zhou, X.~Han, Z.~Hu, and Y.~Ji, ``Knowledge graph embedding for hyper-relational data,'' \emph{Tsinghua Science and Technology}, vol.~22, no.~2, pp. 185--197, 2017.

\bibitem{Xiao2015TransA}
H.~Xiao, M.~Huang, Y.~Hao, and X.~Zhu, ``Trans{A}: An adaptive approach for knowledge graph embedding,'' \emph{CoRR}, vol. abs/1509.05490, 2015.

\bibitem{Fan2014TransitionbasedKG}
M.~Fan, Q.~Zhou, E.~Chang, and T.~F. Zheng, ``Transition-based knowledge graph embedding with relational mapping properties,'' in \emph{Proceedings of Twenty-Eighth Pacific Asia Conference on Language, Information, and Computing}, 2014, pp. 328--337.

\bibitem{Feng2016TransF}
J.~Feng, M.~Huang, M.~Wang, M.~Zhou, Y.~Hao, and X.~Zhu, ``Knowledge graph embedding by flexible translation,'' in \emph{Proceedings of Fifteenth International Conference on Principles of Knowledge Representation and Reasoning}, 2016, pp. 557--560.

\bibitem{xie2017DomainNS}
Q.~Xie, X.~Ma, Z.~Dai, and E.~Hovy, ``An interpretable knowledge transfer model for knowledge base completion,'' in \emph{Proceedings of Annual Meeting of the Association for Computational Linguistics}, 2017, pp. 950--962.

\bibitem{Qian2018TransAt}
W.~Qian, C.~Fu, Y.~Zhu, D.~Cai, and X.~He, ``Translating embeddings for knowledge graph completion with relation attention mechanism,'' in \emph{Proceedings of Twenty-Seventh International Joint Conference on Artificial Intelligence}, 2018, pp. 4286--4292.

\bibitem{Yang2019TransMS}
S.~Yang, J.~Tian, H.~Zhang, J.~Yan, H.~He, and Y.~Jin, ``Trans{MS}: Knowledge graph embedding for complex relations by multidirectional semantics,'' in \emph{Proceedings of Twenty-Eighth International Joint Conference on Artificial Intelligence}, 2019, pp. 1935--1942.

\bibitem{Xiao2016ManifoldE}
H.~Xiao, M.~Huang, and X.~Zhu, ``From one point to a manifold: Knowledge graph embedding for precise link prediction,'' in \emph{Proceedings of Twenty-Fifth International Joint Conference on Artificial Intelligence}, 2016, pp. 1315--1321.

\bibitem{He2015K2GE}
S.~He, K.~Liu, G.~Ji, and J.~Zhao, ``Learning to represent knowledge graphs with gaussian embedding,'' in \emph{Proceedings of Twenty-Fourth ACM International on Conference on Information and Knowledge Management}, 2015, pp. 623--632.

\bibitem{Xiao2016Transg}
H.~Xiao, M.~Huang, and X.~Zhu, ``Trans{G} : {A} generative model for knowledge graph embedding,'' in \emph{Proceedings of Fifty-Fourth Annual Meeting of the Association for Computational Linguistics}, 2016, pp. 2316--2325.

\bibitem{Zhang2020HAKE}
Z.~Zhang, J.~Cai, Y.~Zhang, and J.~Wang, ``Learning hierarchy-aware knowledge graph embeddings for link prediction,'' in \emph{Proceedings of Thirty-Fourth AAAI Conference on Artificial Intelligence, Innovative Applications of Artificial Intelligence Conference, and Symposium on Educational Advances in Artificial Intelligence}, 2020, pp. 3065--3072.

\bibitem{Takuma2018TorusE}
T.~Ebisu and R.~Ichise, ``Torus{E}: Knowledge graph embedding on a lie group,'' in \emph{Proceedings of the Thirty-Second AAAI Conference on Artificial Intelligence and Thirtieth Innovative Applications of Artificial Intelligence Conference and Eighth AAAI Symposium on Educational Advances in Artificial}, 2018, pp. 1819--1826.

\bibitem{Zhang2019QuatE}
S.~Zhang, Y.~Tay, L.~Yao, and Q.~Liu, ``Quaternion knowledge graph embeddings,'' \emph{Advances in neural information processing systems}, vol.~32, p. 246, 2019.

\bibitem{Sun2019RotatEKG}
Z.~Sun, Z.~Deng, J.~Nie, and J.~Tang, ``Rotat{E}: Knowledge graph embedding by relational rotation in complex space,'' in \emph{Proceedings of International Conference on Learning Representations}, 2019.

\bibitem{Chen2021MobiusE}
Y.~Chen, J.~Liu, Z.~Zhang, S.~Wen, and W.~Xiong, ``M\"{o}{B}ius{E}: Knowledge graph embedding on m\"{o}bius ring,'' \emph{Knowledge Based Systems}, vol. 227, p. 107181, 2021.

\bibitem{Nickel2011ATM}
M.~Nickel, V.~Tresp, and H.~Kriegel, ``A three-way model for collective learning on multi-relational data,'' in \emph{Proceedings of Twenty-Eighth International Conference on Machine Learning}, 2011, pp. 809--816.

\bibitem{Yang2015EmbeddingEA}
B.~Yang, W.~Yih, X.~He, J.~Gao, and L.~Deng, ``Embedding entities and relations for learning and inference in knowledge bases,'' in \emph{Proceedings of International Conference on Learning Representations}, 2015.

\bibitem{Nickel2016HolographicEO}
M.~Nickel, L.~Rosasco, and T.~A. Poggio, ``Holographic embeddings of knowledge graphs,'' in \emph{Proceedings of Thirtieth AAAI Conference on Artificial Intelligence}, 2016, pp. 1955--1961.

\bibitem{Trouillon2016ComplexEF}
T.~Trouillon, J.~Welbl, S.~Riedel, {\'{E}}.~Gaussier, and G.~Bouchard, ``Complex embeddings for simple link prediction,'' in \emph{Proceedings of Thirty-Third International Conference on Machine Learning}, 2016, pp. 2071--2080.

\bibitem{Kazemi2018SimplEEF}
S.~M. Kazemi and D.~Poole, ``Simple embedding for link prediction in knowledge graphs,'' in \emph{Proceedings of Thirty-Second International Conference on Neural Information Processing Systems}, 2018, pp. 4289--4300.

\bibitem{Xue2018Holex}
Y.~Xue, Y.~Yuan, Z.~Xu, and A.~Sabharwal, ``Expanding holographic embeddings for knowledge completion,'' in \emph{Proceedings of Thirty-Second International Conference on Neural Information Processing Systems}, 2018, pp. 4496--4506.

\bibitem{Zhang2019CrossE}
W.~Zhang, B.~Paudel, W.~Zhang, A.~Bernstein, and H.~Chen, ``Interaction embeddings for prediction and explanation in knowledge graphs,'' in \emph{Proceedings of Twelfth ACM International Conference on Web Search and Data Mining}, 2019, pp. 96--104.

\bibitem{Liu2017Analogical}
H.~Liu, Y.~Wu, and Y.~Yang, ``Analogical inference for multi-relational embeddings,'' in \emph{Proceedings of Thirty-Fourth International Conference on Machine Learning}, 2017, pp. 2168--2178.

\bibitem{Balazevic2019TuckERTF}
I.~Balazevic, C.~Allen, and T.~M. Hospedales, ``Tuck{ER}: Tensor factorization for knowledge graph completion,'' in \emph{Proceedings of Conference on Empirical Methods in Natural Language Processing and Ninth International Joint Conference on Natural Language Processing}, 2019, pp. 5184--5193.

\bibitem{Dettmers2018Convolutional2K}
T.~Dettmers, P.~Minervini, P.~Stenetorp, and S.~Riedel, ``Convolutional 2d knowledge graph embeddings,'' in \emph{Proceedings of the Thirty-Second AAAI Conference on Artificial Intelligence and Thirtieth Innovative Applications of Artificial Intelligence Conference and Eighth AAAI Symposium on Educational Advances in Artificial Intelligence}, 2018, pp. 1811--1818.

\bibitem{Nguyen2018Convkb}
D.~Q. Nguyen, T.~D. Nguyen, D.~Q. Nguyen, and D.~Phung, ``A novel embedding model for knowledge base completion based on convolutional neural network,'' in \emph{Proceedings of the Conference of the North {A}merican Chapter of the Association for Computational Linguistics}, 2018, pp. 327--333.

\bibitem{Jiang2019ConvR}
X.~Jiang, Q.~Wang, and B.~Wang, ``Adaptive convolution for multi-relational learning,'' in \emph{Proceedings of the Conference of the North {A}merican Chapter of the Association for Computational Linguistics}, 2019, pp. 978--987.

\bibitem{Shikhar2020interacte}
S.~Vashishth, S.~Sanyal, V.~Nitin, N.~Agrawal, and P.~Talukdar, ``Interacte: Improving convolution-based knowledge graph embeddings by increasing feature interactions,'' in \emph{Proceedings of the AAAI Conference on Artificial Intelligence}.\hskip 1em plus 0.5em minus 0.4em\relax AAAI Press, 2020, pp. 3009--3016.

\bibitem{Nguyen2019CapsE}
D.~Q. Nguyen, T.~Vu, T.~D. Nguyen, D.~Q. Nguyen, and D.~Phung, ``A capsule network-based embedding model for knowledge graph completion and search personalization,'' in \emph{Proceedings of the Association for Computational Linguistics}, 2019, pp. 2180--2189.

\bibitem{Yao2019KGBERTBF}
L.~Yao, C.~Mao, and Y.~Luo, ``{KG-BERT:} {BERT} for knowledge graph completion,'' \emph{CoRR}, vol. abs/1909.03193, 2019.

\bibitem{Devlin2019BERT}
J.~Devlin, M.-W. Chang, K.~Lee \emph{et~al.}, ``{BERT}: Pre-training of deep bidirectional transformers for language understanding,'' in \emph{Proceedings of Conference of the North {A}merican Chapter of the Association for Computational Linguistics: Human Language Technologies}, 2019, pp. 4171--4186.

\bibitem{Zhang2019Ernie}
Z.~Zhang, X.~Han, Z.~Liu, X.~Jiang, M.~Sun, and Q.~Liu, ``{ERNIE}: Enhanced language representation with informative entities,'' in \emph{Proceedings of the Annual Meeting of the Association for Computational Linguistics}, 2019, pp. 1441--1451.

\bibitem{Bo2021Star}
B.~Wang, T.~Shen, G.~Long, T.~Zhou, Y.~Wang, and Y.~Chang, ``Structure-augmented text representation learning for efficient knowledge graph completion,'' in \emph{Proceedings of the Web Conference}, 2021, pp. 1737--1748.

\bibitem{Wang2022Simkgc}
L.~Wang, W.~Zhao, Z.~Wei, and J.~Liu, ``{S}im{KGC}: Simple contrastive knowledge graph completion with pre-trained language models,'' in \emph{Proceedings of the Annual Meeting of the Association for Computational Linguistics}, 2022, pp. 4281--4294.

\bibitem{Wang2019KEPLERAU}
X.~Wang, T.~Gao, Z.~Zhu, Z.~Liu, J.-Z. Li, and J.~Tang, ``Kepler: A unified model for knowledge embedding and pre-trained language representation,'' \emph{Transactions of the Association for Computational Linguistics}, vol.~9, pp. 176--194, 2019.

\bibitem{Wang2021InductivE}
B.~Wang, G.~Wang, J.~Huang, J.~You, J.~Leskovec, and C.-C.~J. Kuo, ``Inductive learning on commonsense knowledge graph completion,'' in \emph{Proceedings of the International Joint Conference on Neural Networks}, 2021, pp. 1--8.

\bibitem{Zha2022InductiveRP}
H.~Zha, Z.~Chen, and X.~Yan, ``Inductive relation prediction by bert,'' in \emph{Proceedings of the AAAI Conference on Artificial Intelligence}, vol.~36, 2022, pp. 5923--5931.

\bibitem{saxena2022kgt5}
A.~Saxena, A.~Kochsiek, and R.~Gemulla, ``Sequence-to-sequence knowledge graph completion and question answering,'' in \emph{Proceedings of the Annual Meeting of the Association for Computational Linguistics}, 2022, pp. 2814--2828.

\bibitem{Schroff2015FaceNet}
F.~Schroff, D.~Kalenichenko, and J.~Philbin, ``Facenet: A unified embedding for face recognition and clustering,'' in \emph{Proceedings of IEEE Conference on Computer Vision and Pattern Recognition}, 2015, pp. 815--823.

\bibitem{Gutmann2012NCE}
M.~U. Gutmann and A.~Hyv\"{a}rinen, ``Noise-contrastive estimation of unnormalized statistical models, with applications to natural image statistics,'' \emph{The Journal of Machine Learning Research}, vol.~13, pp. 307--361, 2012.

\bibitem{Gutmann2010Noise}
M.~Gutmann and A.~Hyv{\"a}rinen, ``Noise-contrastive estimation: A new estimation principle for unnormalized statistical models,'' in \emph{Proceedings of International Conference on Artificial Intelligence and Statistics}, 2010, pp. 297--304.

\bibitem{Mikolov2013DistRepresent}
T.~Mikolov, I.~Sutskever, K.~Chen, G.~Corrado, and J.~Dean, ``Distributed representations of words and phrases and their compositionality,'' in \emph{Proceedings of International Conference on Neural Information Processing Systems}, 2013, pp. 3111--3119.

\bibitem{Mnih2012NCE}
A.~Mnih and Y.~W. Teh, ``A fast and simple algorithm for training neural probabilistic language models,'' in \emph{Proceedings of Twenty-Ninth International Conference on International Conference on Machine Learning}, 2012, pp. 419--426.

\bibitem{Reiter1978OWS}
R.~Reiter, \emph{Deductive Question-Answering on Relational Data Bases}, 1978, ch. Logic and Data Bases, pp. 149--177.

\bibitem{Yang2020underneg}
Z.~Yang, M.~Ding, C.~Zhou, H.~Yang, J.~Zhou, and J.~Tang, ``Understanding negative sampling in graph representation learning,'' in \emph{Proceedings of Twenty-Sixth Conference on Knowledge Discovery and Data Mining}, 2020, pp. 1666--1676.

\bibitem{Peng2018RandomCorrupt}
P.~Xu and D.~Barbosa, ``Investigations on knowledge base embedding for relation prediction and extraction,'' \emph{CoRR}, vol. abs/1802.02114, 2018.

\bibitem{Lerer2019BatchNS}
A.~Lerer, L.~Wu, J.~Shen, T.~Lacroix, L.~Wehrstedt, A.~Bose, and A.~Peysakhovich, ``{Pytorch-BigGraph}: A large scale graph embedding system,'' in \emph{Proceedings of The Conference on Machine Learning and Systems}, 2019, pp. 120--131.

\bibitem{Kanojia2017pns}
V.~Kanojia, H.~Maeda, R.~Togashi, and S.~Fujita, ``Enhancing knowledge graph embedding with probabilistic negative sampling,'' in \emph{Proceedings of Twenty-sixth International Conference on World Wide Web Companion}, 2017, pp. 801--802.

\bibitem{Zhang2019MYX}
Y.~Zhang, W.~Cao, and J.~Liu, ``A novel negative sample generating method for knowledge graph embedding,'' in \emph{Proceedings of International Conference on Embedded Wireless Systems and Networks}, 2019, pp. 401--406.

\bibitem{Cao2021SparseNSG}
W.~Cao, Y.~Zhang, J.~Liu, and Z.~Rao, ``A novel negative sampling based on frequency of relational association entities for knowledge graph embedding,'' \emph{Journal of Web Engineering}, vol.~20, no.~6, pp. 1867--1884, 2021.

\bibitem{Yao2023ERDNS}
N.~Yao, Q.~Liu, Y.~Yang, W.~Li, and Q.~Bai, ``Entity-relation distribution-aware negative sampling for knowledge graph embedding,'' in \emph{Proceedings of the International Semantic Web Conference}, 2023, pp. 234--252.

\bibitem{Lijuan2024GNS}
L.~Duan, S.~Han, W.~Jiang, M.~He, and Y.~Qiao, ``Link prediction based on data augmentation and metric learning knowledge graph embedding,'' \emph{Applied Sciences}, vol.~14, no.~8, 2024.

\bibitem{kotnis2017analysis}
B.~Kotnis and V.~Nastase, ``Analysis of the impact of negative sampling on link prediction in knowledge graphs,'' \emph{CoRR}, vol. abs/1708.06816, 2017.

\bibitem{Alam2022Lemon}
M.~M. Alam, M.~R. A.~H. Rony, S.~Ali, J.~Lehmann, and S.~Vahdati, ``Language model-driven negative sampling,'' \emph{CoRR}, vol. abs/2203.04703, 2022.

\bibitem{Tian2022LTS}
W.~Tian and L.~Pan, ``Knowledge alignment based on negative sampling of potential topic similarities,'' in \emph{Proceedings of International Conference on Image Processing, Computer Vision and Machine Learning}, 2022, pp. 249--253.

\bibitem{Xie2016STC}
R.~Xie, Z.~Liu, and M.~Sun, ``Representation learning of knowledge graphs with hierarchical types,'' in \emph{Proceedings of the International Joint Conference on Artificial Intelligence}, 2016, pp. 2965--2971.

\bibitem{Wang2022KGBoost}
Y.-C. Wang, X.~Ge, B.~Wang, and C.-C.~J. Kuo, ``{KGB}oost: A classification-based knowledge base completion method with negative sampling,'' \emph{Pattern Recognition Letters}, vol. 157, pp. 104--111, 2022.

\bibitem{Juan2019Transx}
J.~Li, W.~Zhang, and H.~Chen, ``Incorporating domain and range of relations for knowledge graph completion,'' in \emph{Proceedings of China Conference on Knowledge Graph and Semantic Computing}, 2019, pp. 50--61.

\bibitem{Weyns2020ConditionalCF}
M.~Weyns, P.~Bonte, B.~Steenwinckel, F.~D. Turck, and F.~Ongenae, ``Conditional constraints for knowledge graph embeddings,'' in \emph{Proceedings of Workshop on Deep Learning for Knowledge Graphs}, 2020.

\bibitem{Ahrabian2020StructureAN}
K.~Ahrabian, A.~Feizi, Y.~Salehi, W.~L. Hamilton, and A.~J. Bose, ``Structure aware negative sampling in knowledge graphs,'' in \emph{Proceedings of Conference on Empirical Methods in Natural Language Processing}, 2020, pp. 6093--6101.

\bibitem{Zhu2023Adaptive}
D.~Zhu, H.~Tan, L.~Wang, Y.~Feng, Y.~Lin, and Z.~Gu, ``Knowledge graph embedding via adaptive negative subsampling,'' in \emph{Proceeding of the International Conference on Data Science in Cyberspace}, 2023, pp. 38--43.

\bibitem{Wang2019CDNS}
Y.~Wang, Y.~Liu, H.~Zhang, and H.~Xie, ``Leveraging lexical semantic information for learning concept-based multiple embedding representations for knowledge graph completion,'' in \emph{Proceedings of International Conference on Web and Big Data}, 2019, pp. 382--397.

\bibitem{Yuan2023ConceptDriven}
L.~Yuan, Y.~Wang, T.~Chen, X.~Zhang, C.~Liu, and Y.~Zhang, ``A novel concept-driven negative sampling mechanism for enhancing semanticity and interpretability of knowledge graph completion task,'' in \emph{Proceedings of the International Conference on Data Science in Cyberspace}, 2023, pp. 277--284.

\bibitem{lopez2023GNS}
Y.~A. L\'{o}pez-Rodr\'{\i}guez, O.~G. Toledano-L\'{o}pez, Y.~Hidalgo-Delgado, H.~Gonz\'{a}lez~Di\'{e}z, and R.~Segundo-Guerrero, ``Good negative sampling for\&nbsp;triple classification,'' in \emph{Proceedings of the International Congress on Artificial Intelligence and Pattern Recognition}, 2023, pp. 323--334.

\bibitem{Qin2019ANS}
S.~Qin, G.~Rao, C.~Bin, L.~Chang, T.~Gu, and W.~Xuan, ``Knowledge graph embedding based on adaptive negative sampling,'' in \emph{Proceedings of International Conference of Pioneering Computer Scientists, Engineers, and Educators}, 2019, pp. 551--563.

\bibitem{Je2022EntityAN}
S.~hyun Je, ``Entity aware negative sampling with auxiliary loss of false negative prediction for knowledge graph embedding,'' \emph{CoRR}, vol. abs/2210.06242, 2022.

\bibitem{lin2023HTENS}
Z.~Lin, Z.~Zhao, J.~Xie, and Y.~Shen, ``Hierarchical type enhanced negative sampling for knowledge graph embedding,'' in \emph{Proceedings of the International ACM SIGIR Conference on Research and Development in Information Retrieval}, 2023, pp. 2047--2051.

\bibitem{Peng2023TypeAugment}
P.~He, G.~Zhou, Y.~Yao, Z.~Wang, and H.~Yang, ``A type-augmented knowledge graph embedding framework for knowledge graph completion,'' \emph{Scientific Reports}, vol.~13, no. 12364, 2023.

\bibitem{Sun2018Bootstrapping}
Z.~Sun, W.~Hu, Q.~Zhang, and Y.~Qu, ``Bootstrapping entity alignment with knowledge graph embedding,'' in \emph{Proceedings of Twenty-Seventh International Joint Conference on Artificial Intelligence}, 2018, pp. 4396--4402.

\bibitem{Truncated2023Li}
H.~Li, Z.~Zhu, H.~Zhu, and B.~Jin, ``Fusing attribute character embeddings with truncated negative sampling for entity alignment,'' \emph{Electronics}, vol.~12, no.~8, p. 1947, 2023.

\bibitem{Alam2020AffinityDN}
M.~M. Alam, H.~Jabeen, M.~Ali, K.~Mohiuddin, and J.~Lehmann, ``Affinity dependent negative sampling for knowledge graph embeddings,'' in \emph{Proceedings of Workshop on Deep Learning for Knowledge Graphs}, 2020.

\bibitem{Dash2019DNS}
S.~Dash and A.~Gliozzo, ``Distributional negative sampling for knowledge base completion,'' \emph{CoRR}, vol. abs/1908.06178, 2019.

\bibitem{Che2022MixKGMF}
F.~Che, G.~Yang, P.~Shao, D.~Zhang, and J.~Tao, ``{MixKG}: Mixing for harder negative samples in knowledge graph,'' \emph{CoRR}, vol. abs/2202.09606, 2022.

\bibitem{Islam2022SNS}
M.~K. Islam, S.~Aridhi, and M.~Smail-Tabbone, ``Simple negative sampling for link prediction in knowledge graphs,'' in \emph{Proceedings of International Conference on Complex Networks and Their Applications}, 2022, pp. 549--562.

\bibitem{Nie2023DSSS}
H.~Nie, X.~Zhao, X.~Bi, Y.~Ma, and G.~Y. Yuan, ``Correlation embedding learning with dynamic semantic enhanced sampling for knowledge graph completion,'' \emph{World Wide Web}, vol.~26, no.~5, pp. 2887--2907, 2023.

\bibitem{Jain2021ReasonKGE}
N.~Jain, T.-K. Tran, M.~H. Gad-Elrab, and D.~Stepanova, ``Improving knowledge graph embeddings with ontological reasoning,'' in \emph{Proceedings of International Semantic Web Conference}, 2021, pp. 410--426.

\bibitem{Cai2018KBGANAL}
L.~Cai and W.~Y. Wang, ``{KBGAN:} adversarial learning for knowledge graph embeddings,'' in \emph{Proceedings of Conference of the North {A}merican Chapter of the Association for Computational Linguistics: Human Language Technologies}, 2018, pp. 1470--1480.

\bibitem{Wang2018IncorporatingGF}
P.~Wang, S.~Li, and R.~Pan, ``Incorporating {GAN} for negative sampling in knowledge representation learning,'' in \emph{Proceedings of Thirty-Second AAAI Conference on Artificial Intelligence}, 2018, pp. 2005--2012.

\bibitem{Wang2018GraphGAN}
H.~Wang, J.~Wang, J.~Wang, M.~Zhao, W.~Zhang, F.~Zhang, X.~Xie, and M.~Guo, ``{GraphGAN}: Graph representation learning with generative adversarial nets,'' in \emph{Proceedings of the Thirty-Second AAAI Conference on Artificial Intelligence and Thirtieth Innovative Applications of Artificial Intelligence Conference and Eighth AAAI Symposium on Educational Advances in Artificial Intelligence}, 2018, pp. 2508--2515.

\bibitem{Zia2021KCGAN}
T.~Zia and D.~Windridge, ``A generative adversarial network for single and multi-hop distributional knowledge base completion,'' \emph{Neurocomputing}, pp. 543--551, 2021.

\bibitem{LIU2022GNDN}
L.~Liu, J.~Zeng, and X.~Zheng, ``Learning structured embeddings of knowledge graphs with generative adversarial framework,'' \emph{Expert Systems with Applications}, vol. 204, p. 117361, 2022.

\bibitem{Kairong2019KSGAN}
K.~Hu, H.~Liu, and T.~Hao, ``A knowledge selective adversarial network for link prediction in knowledge graph,'' in \emph{Proceedings of Conference on Natural Language Processing and Chinese Computing}, 2019, pp. 171--183.

\bibitem{Hai2020NKSGAN}
H.~Liu, K.~Hu, F.~L. Wang, and T.~Hao, ``Aggregating neighborhood information for negative sampling for knowledge graph embedding,'' \emph{Neural Computing and Applications}, vol.~32, no.~23, pp. 17\,637--17\,653, 2020.

\bibitem{Caifang2021RUGA}
T.~Caifang, R.~Yuan, Y.~Hualei, S.~Ling, C.~Jiamin, and W.~Yutian, ``Improving knowledge graph completion using soft rules and adversarial learning,'' \emph{Chinese Journal of Electronics}, vol.~30, no.~4, pp. 623--633, 2021.

\bibitem{Cheng2020NoiGAN}
K.~Cheng, Y.~Zhu, M.~Zhang, and Y.~Sun, ``Noigan: Noise aware knowledge graph embedding with adversarial learning,'' 2020.

\bibitem{Le2021NAGAN-ConvKB}
T.~Le, T.~Pham, and B.~Le, ``Negative sampling for knowledge graph completion based on generative adversarial network,'' in \emph{Proceedings of International Conference on Computational Collective Intelligence}, 2021, pp. 3--15.

\bibitem{Shan2018ConfidenceAware}
Y.~Shan, C.~Bu, X.~Liu, S.~Ji, and L.~Li, ``Confidence-aware negative sampling method for noisy knowledge graph embedding,'' in \emph{Proceedings of IEEE International Conference on Big Knowledge}, 2018, pp. 33--40.

\bibitem{Zhang2019NSCachingSA}
Y.~Zhang, Q.~Yao, Y.~Shao, and L.~Chen, ``{NSC}aching: Simple and efficient negative sampling for knowledge graph embedding,'' in \emph{Proceedings of IEEE International Conference on Data Engineering}, 2019, pp. 614--625.

\bibitem{Lei2019LAS}
J.~Lei, D.~Ouyang, and Y.~Liu, ``Adversarial knowledge representation learning without external model,'' \emph{IEEE Access}, vol.~7, pp. 3512--3524, 2019.

\bibitem{qin2021ASA}
X.~Qin, N.~Sheikh, B.~Reinwald, and L.~Wu, ``Relation-aware graph attention model with adaptive self-adversarial training,'' in \emph{Proceedings of Thirty-Fifth AAAI Conference on Artificial Intelligence}, 2021, pp. 9368--9376.

\bibitem{Yao2022ESNS}
N.~Yao, Q.~Liu, X.~Li, Y.~Yang, and Q.~Bai, ``Entity similarity-based negative sampling for knowledge graph embedding,'' in \emph{Proceedings of Pacific Rim International Conference on Artificial Intelligence}, 2022, pp. 73--87.

\bibitem{si2023attention}
S.~Cen, X.~Wang, X.~Zou, C.~Liu, and G.~Dai, ``New attention strategy for negative sampling in knowledge graph embedding,'' \emph{Applied Intelligence}, vol.~53, pp. 26\,418--26\,438, 2023.

\bibitem{Feng2024TANS}
X.~Feng, H.~Kamigaito, K.~Hayashi, and T.~Watanabe, ``Unified interpretation of smoothing methods for negative sampling loss functions in knowledge graph embedding,'' \emph{CoRR}, vol. abs/2407.04251, 2024.

\bibitem{tiroshan2022mdncaching}
T.~Madushanka and R.~Ichise, ``{MDN}caching: A strategy to~generate quality negatives for~knowledge graph embedding,'' in \emph{Proceedings of International Conference on Industrial, Engineering and Other Applications of Applied Intelligent Systems}, 2022, pp. 877--888.

\bibitem{Han2023CCS}
H.~Han, X.~Li, and K.~Wu, ``{Op-Trans}: An optimization framework for negative sampling and triplet-mapping properties in knowledge graph embedding,'' \emph{Applied Sciences}, vol.~13, no.~5, p. 2817, 2023.

\bibitem{Zhang2024HaSa}
H.~Zhang, J.~Zhang, and I.~Molybog, ``Hasa: Hardness and structure-aware contrastive knowledge graph embedding,'' in \emph{Proceedings of the ACM Web Conference}, 2024, pp. 2116--2127.

\bibitem{huang2020rate}
H.~Huang, G.~Long, T.~Shen, J.~Jiang, and C.~Zhang, ``{R}at{E}: Relation-adaptive translating embedding for knowledge graph completion,'' in \emph{Proceedings of Twenty-Eighth International Conference on Computational Linguistics}, 2020, pp. 556--567.

\bibitem{Niu2022cake}
G.~Niu, B.~Li, Y.~Zhang, and S.~Pu, ``{CAKE}: A scalable commonsense-aware framework for multi-view knowledge graph completion,'' in \emph{Proceedings of the Annual Meeting of the Association for Computational Linguistics}, 2022, pp. 2867--2877.

\bibitem{Chen2023DeMix}
X.~Chen, W.~Zhang, Z.~Yao, M.~Chen, and S.~Tang, ``Negative sampling with~adaptive denoising mixup for~knowledge graph embedding,'' in \emph{Proceedings of International Semantic Web Conference}, 2023, pp. 253--270.

\bibitem{Che2024M2ixKG}
F.~Che and J.~Tao, ``M2ixkg: Mixing for harder negative samples in knowledge graph,'' \emph{Neural Networks}, vol. 177, p. 106358, 2024.

\bibitem{Zheng2024DCNS}
H.~Zheng, D.~Guan, S.~Xu, and W.~Yuan, ``Dcns: A double-cache negative sampling method for improving knowledge graph embedding,'' in \emph{Proceedings of the International Joint Conference of Web and Big Data}, 2024, pp. 438--450.

\bibitem{Qiao2023GHN}
Z.~Qiao, W.~Ye, D.~Yu, T.~Mo, W.~Li, and S.~Zhang, ``Improving knowledge graph completion with generative hard negative mining,'' in \emph{Proceedings of the Conference of the Association for Computational Linguistics}, 2023, pp. 5866--5878.

\bibitem{Reimers2019SentenceBERTSE}
N.~Reimers and I.~Gurevych, ``{Sentence-BERT}: Sentence embeddings using siamese bert-networks,'' in \emph{Proceedings of the Conference on Empirical Methods in Natural Language Processing}, 2019, pp. 8328--8350.

\bibitem{Hervé2010PCA}
H.~Abdi and L.~J. Williams, ``Principal component analysis,'' \emph{Wiley Interdisciplinary Reviews: Computational Statistics}, vol.~2, no.~4, pp. 433--459, 2010.

\bibitem{David2007kmeans}
D.~Arthur and S.~Vassilvitskii, ``k-means++: the advantages of careful seeding,'' in \emph{Proceedings of Annual ACM-SIAM symposium on Discrete algorithms}, 2007, pp. 1027--1035.

\bibitem{Wu2012Probase}
W.~Wu, H.~Li, H.~Wang, and K.~Q. Zhu, ``Probase: a probabilistic taxonomy for text understanding,'' in \emph{Proceedings of the ACM SIGMOD International Conference on Management of Data}, 2012, pp. 481--492.

\bibitem{Hartigan1979KMeans}
J.~A. Hartigan and M.~A. Wong, ``Algorithm {AS} 136: A {K-Means} clustering algorithm,'' \emph{Applied Statistics}, vol.~28, no.~1, pp. 100--108, 1979.

\bibitem{Goodfellow2014GenerativeAN}
I.~J. Goodfellow, J.~Pouget{-}Abadie, M.~Mirza, B.~Xu, D.~Warde{-}Farley, S.~Ozair, A.~C. Courville, and Y.~Bengio, ``Generative adversarial nets,'' in \emph{Proceedings of International Conference on Neural Information Processing Systems}, 2014, pp. 2672--2680.

\bibitem{Hongyi2018Mixup}
H.~Zhang, M.~Ciss{\'{e}}, Y.~N. Dauphin, and D.~Lopez{-}Paz, ``mixup: Beyond empirical risk minimization,'' in \emph{Proceedings of Sixth International Conference on Learning Representations}, 2018.

\bibitem{Lewis2020Bart}
M.~Lewis, Y.~Liu, N.~Goyal, M.~Ghazvininejad, A.~Mohamed, O.~Levy, V.~Stoyanov, and L.~Zettlemoyer, ``{BART}: Denoising sequence-to-sequence pre-training for natural language generation, translation, and comprehension,'' in \emph{Proceedings of Annual Meeting of the Association for Computational Linguistics}, 2020, pp. 7871--7880.

\bibitem{Seon2022BootstrappedKGE}
J.~S. Kim, S.~J. Ahn, and M.~H. Kim, ``Bootstrapped knowledge graph embedding based on neighbor expansion,'' in \emph{Proceedings of Thirty-First ACM International Conference on Information and Knowledge Management}, 2022, pp. 4123--4127.

\bibitem{hajimoradlou2022stay}
A.~Hajimoradlou and M.~Kazemi, ``Stay positive: knowledge graph embedding without negative sampling,'' \emph{CoRR}, vol. abs/2201.02661, 2022.

\bibitem{Zelong2021NSKGE}
Z.~Li, J.~Ji, Z.~Fu, Y.~Ge, S.~Xu, C.~Chen, and Y.~Zhang, ``Efficient non-sampling knowledge graph embedding,'' in \emph{Proceedings of the Web Conference}, 2021, pp. 1727--1736.

\bibitem{Bahaj2024NSFKGE}
B.~Adil and G.~Mounir, ``Negative-sample-free knowledge graph embedding,'' \emph{Data Mining and Knowledge Discovery}, 2024.

\bibitem{bahaj2022kgnsf}
A.~Bahaj, S.~Lhazmir, and M.~Ghogho, ``Kg-nsf: Knowledge graph completion with a negative-sample-free approach,'' \emph{CoRR}, vol. abs/2207.14617, 2022.

\bibitem{Jure2021BTLoss}
J.~Zbontar, L.~Jing, I.~Misra, Y.~LeCun, and S.~Deny, ``Barlow twins: Self-supervised learning via redundancy reduction,'' in \emph{Proceedings of the International Conference on Machine Learning}, vol. 139, 2021, pp. 12\,310--12\,320.

\end{thebibliography}
%



\end{document}